\documentclass{article} 
\usepackage{iclr2023_conference,times}

\usepackage{amsmath,amsfonts,bm}

\def\eqref#1{equation~\ref{#1}}

\def\1{\bm{1}}

\def\vs{{\bm{s}}}

\DeclareMathAlphabet{\mathsfit}{\encodingdefault}{\sfdefault}{m}{sl}
\SetMathAlphabet{\mathsfit}{bold}{\encodingdefault}{\sfdefault}{bx}{n}

\usepackage[utf8]{inputenc} 
\usepackage[T1]{fontenc}    
\usepackage{hyperref}       
\usepackage{url}            
\usepackage{booktabs}       
\usepackage{amsfonts}       
\usepackage{nicefrac}       
\usepackage{microtype}      
\usepackage{xcolor}         
\usepackage{color}
\usepackage{mathrsfs}
\usepackage{amsmath}
\usepackage{graphicx}
\usepackage{multirow}
\usepackage{amssymb}
\usepackage{makecell}

\usepackage[symbol]{footmisc} 
\usepackage{float}  

\usepackage{soul} 

\usepackage{colortbl}
\usepackage{adjustbox}
\usepackage{tabularx}
\usepackage{tabu}
\usepackage{caption}

\usepackage{pifont}  
\newcommand{\xmark}{\ding{55}}%
\newcommand{\vmark}{\ding{51}}%

\definecolor{Gray}{gray}{0.93}
\definecolor{Cyan}{rgb}{0.88,1,1}
\usepackage{soul}
\sethlcolor{Gray}

\newcommand{\PreserveBackslash}[1]{\let\temp=\\#1\let\\=\temp}
\newcolumntype{C}[1]{>{\PreserveBackslash\centering}p{#1}}
\newcolumntype{L}[1]{>{\PreserveBackslash\raggedright}p{#1}}
\newcolumntype{R}[1]{>{\PreserveBackslash\raggedleft}p{#1}}

\definecolor{convcolor}{HTML}{412F8A}
\definecolor{resnetcolor}{HTML}{8DA0CB}
\definecolor{vitcolor}{HTML}{fc8e62}
\newcommand{\convcolor}[1]{\textcolor{convcolor}{#1}}
\newcommand{\vitcolor}[1]{\textcolor{vitcolor}{#1}}
\newcommand{\bv}{\vitcolor{$\mathbf{\circ}$\,}}
\newcommand{\bc}{\convcolor{$\bullet$\,}}
\newcommand{\bh}{\vitcolor{$\bullet$\,}}  

\newcommand{\net}[0]{RevCol}
\title{Reversible Column Networks}

\author{Yuxuan Cai\textsuperscript{1} \hspace{1em}  Yizhuang Zhou\textsuperscript{1}\hspace{1em}  Qi Han\textsuperscript{1}\hspace{1em}  Jianjian Sun\textsuperscript{1}\hspace{1em}  Xiangwen Kong\textsuperscript{1}\hspace{1em}  Jun Li\textsuperscript{1}\\
\centerline{\textbf{Xiangyu Zhang\textsuperscript{12} \thanks{Corresponding author. 
 This work is supported by The National Key Research and Development Program of China (No. 2017YFA0700800) and Beijing Academy of Artificial Intelligence (BAAI). }}} \\
\vspace{-5px}
  \\ \centerline{MEGVII Technology\textsuperscript{1}}
  \\ \centerline{Beijing Academy of Artificial Intelligence\textsuperscript{2}}
  \\ \centerline{\texttt{\{caiyuxuan, zhouyizhuang, hanqi, zhangxiangyu\}@megvii.com}}
  }
\usepackage{xspace}

\makeatletter
\DeclareRobustCommand\onedot{\futurelet\@let@token\@onedot}
\def\@onedot{\ifx\@let@token.\else.\null\fi\xspace}

\def\eg{\emph{e.g}\onedot} 
\def\ie{\emph{i.e}\onedot} 
 
 \def\vs{\emph{vs}\onedot}
\def\wrt{w.r.t\onedot} 

\makeatother

\iclrfinalcopy 
\begin{document}

\maketitle
\vspace{-10px}
\begin{abstract}

We propose a new neural network design paradigm \emph{Reversible Column Network (RevCol)}. The main body of RevCol is composed of multiple copies of subnetworks, named \emph{columns} respectively, between which multi-level reversible connections are employed. Such architectural scheme attributes RevCol very different behavior from conventional networks: during forward propagation, features in RevCol are learned to be gradually \emph{disentangled} when passing through each column, whose total information is maintained rather than \emph{compressed} or discarded as other network does. Our experiments suggest that CNN-style RevCol models can achieve very competitive performances on multiple computer vision tasks such as image classification, object detection and semantic segmentation, especially with large parameter budget and large dataset. For example, after ImageNet-22K pre-training, RevCol-XL obtains 88.2\% ImageNet-1K accuracy. Given more pre-training data, our largest model RevCol-H reaches \textbf{90.0\%} on ImageNet-1K, \textbf{63.8\%} AP$_{box}$ on COCO detection minival set, \textbf{61.0\%} mIoU on ADE20k segmentation. To our knowledge, it is the best COCO detection and ADE20k segmentation result among \emph{pure} (static) CNN models. Moreover, as a general macro architecture fashion, RevCol can also be introduced into transformers or other neural networks, which is demonstrated to improve the performances in both computer vision and NLP tasks. 
We release code and models at \url{https://github.com/megvii-research/RevCol}
\end{abstract}

\section{Introduction}
\label{intro}

\begin{figure}[t]
    \centering
    \includegraphics[width=\textwidth]{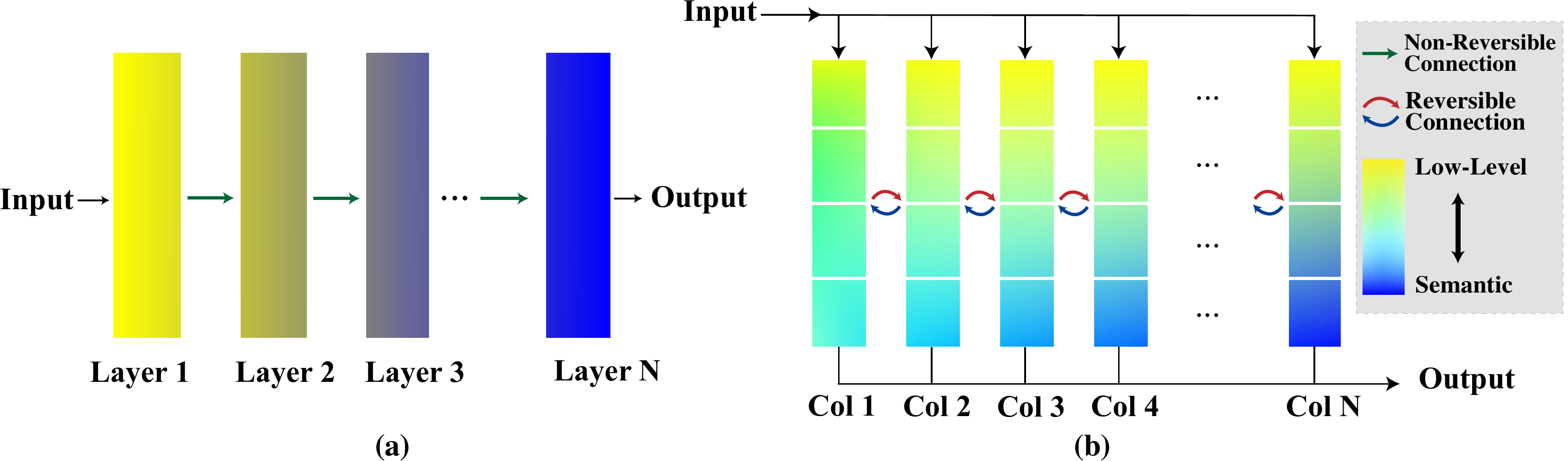}

    \caption{Sketch of the information propagation in: (a) Vanilla single-column network. (b) Our reversible column network. Yellow color denotes low-level information and blue color denotes semantic information. }
    \label{fig:intro}
    \vspace{-10pt}
\end{figure}

\emph{Information Bottleneck principle (IB)} ~\citep{tishby2000information, tishby2015deep} rules the deep learning world. Consider a typical supervised learning network as in Fig.~\ref{fig:intro} (a): layers close to the input contain more low-level information, while features close to the output are rich in semantic meanings. In other words, information unrelated to the target is gradually \emph{compressed} during the layer-by-layer propagation. Although such learning paradigm achieves great success in many practical applications, it might not be the optimal choice in the view of \emph{feature learning} -- down-stream tasks may suffer from inferior performances if the learned features are \emph{over} compressed, or the learned semantic information is irrelevant to the target tasks, especially if a significant domain gap exists between the source and the target tasks \citep{zamir2018taskonomy}. Researchers have devoted great efforts to make the learned features to be more universally applicable, \eg via self-supervised pre-training\citep{oord2018representation,devlin2018bert,he2022masked,xie2022simmim} or multi-task learning \citep{ruder2017overview, caruana1997multitask, sener2018multi}. 

In this paper, we mainly focus on an alternative approach: building a network to learn \emph{disentangled representations}. Unlike \emph{IB} learning, disentangled feature learning \citep{desjardins2012disentangling,bengio2013representation,hinton2021represent} does not intend to extract the most related information while discard the less related; instead, it aims to embed the task-relevant concepts or semantic words into a few decoupled dimensions respectively. 
Meanwhile the whole feature vector roughly maintains as much information as the input. It is quite analogous to 
the mechanism in biological cells \citep{hinton2021represent, lillicrap2020backpropagation} -- each cell shares an identical copy of the whole genome but has different expression intensities. Accordingly in computer vision tasks, learning disentangled features is also reasonable: for instance, high-level semantic representations are tuned during \emph{ImageNet} pre-training, meanwhile the low-level information (\eg locations of the edges) should also be maintained in other feature dimensions in case of the demand of down-stream tasks like object detection.

Fig.~\ref{fig:intro} (b) sketches our main idea: \emph{Reversible Column Networks (RevCol)}, which is greatly inspired by the big picture of \emph{GLOM}~\citep{hinton2021represent}. Our network is composed of $N$ subnetworks (named \emph{columns}) of identical structure (however whose weights are not necessarily the same), each of which receives a copy of the input and generates a prediction. Hence multi-level embeddings, \ie from low-level to highly semantic representations, are stored in each column. Moreover, \emph{reversible transformations} are introduced to propagate the multi-level features from $i$-th column to $(i+1)$-th column without information loss. During the propagation, since the complexity and nonlinearity increases, the quality of all feature levels is expected to gradually improve. Hence the last column (Col $N$ in Fig.~\ref{fig:intro}~(b)) predicts the final disentangled representations of the input. 

In RevCol, one of our key contributions is the design of the \emph{reversible transformations} between adjacent columns. The concept is borrowed from the family of \emph{Reversible Networks} \citep{chang2018reversible,gomez2017reversible,jacobsen2018revnet,mangalam2022reversible}; however, conventional reversible structures such as \emph{RevNets} \citep{gomez2017reversible} (Fig.~\ref{fig:arch} (a)) usually have two drawbacks: first, feature maps within a reversible block are restricted to have the same shape\footnote{In precise, feature maps of odd and even indexes should be equal sized respectively. }; second, the last two feature maps in RevNets have to contain both low-level and high-level information due to the reversible nature, which may be difficult to optimize as in conflict with \emph{IB} principle. In this paper, we overcome the drawbacks by introducing a novel reversible multi-level fusion module. The details are discussed in Sec.~\ref{method}. 

We build a series of CNN-based \emph{RevCol} models under different complexity budgets and evaluate them in mainstream computer vision tasks, such as \emph{ImageNet} classification, \emph{COCO} object detection and instance segmentation, as well as \emph{ADE20K} semantic segmentation. Our models achieve comparable or better results than sophisticated CNNs or \emph{vision transformers} like ConvNeXt \citep{liu2022convnet} and Swin \citep{liu2021swin}. For example, after ImageNet-22K pre-training, our RevCol-XL model obtains \textbf{88.2\%} accuracy on ImageNet-1K without using transformers or large convolutional kernels \citep{ding2022scaling,liu2022convnet,han2021connection}. More importantly, we find RevCol can scale up well to large models and large datasets. Given a larger private pre-training dataset, our biggest model RevCol-H obtains \textbf{90.0\%} accuracy on ImageNet-1K classification, \textbf{63.8\%} $AP_{box}$ on COCO detection minival set, and \textbf{61.0\%} mIoU on ADE20K segmentation, respectively. To our knowledge, it is the best reversible model on those tasks, as well as the best \emph{pure} CNN model on COCO and ADE20K which only involves static kernels without dynamic convolutions \citep{dai2017deformable,ma2020weightnet}. In the appendix, we further demonstrate RevCol can work with transformers \citep{dosovitskiy2020image,devlin2018bert} and get improved results on both computer vision and NLP tasks. Finally, similar to RevNets \citep{gomez2017reversible}, RevCol also shares the bonus of memory saving from the reversible nature, which is particularly important for large model training. 

\paragraph{Relation to previous works. } Although our initial idea on feature disentangling is derived from \emph{GLOM} \citep{hinton2021represent}, in \emph{RevCol} there are a lot of simplifications and modifications. For example, GLOM suggests contrastive auxiliary loss to avoid feature collapse. Contrastive training methods need extra pairs of positive and negative samples, which is complicated and unstable. In RevCol, reversible transformations between columns provides lossless information propagation by nature. As for other multi-scale grid-like architectures such as \emph{HRNets}~\citep{wang2020deep}, \emph{DEQ models} \citep{bai2020multiscale} and \emph{FPNs} \citep{lin2017feature,tan2020efficientdet}, the design purpose of those models is to fuse multi-scale features rather than learn disentangled representations; therefore, in general they still follow the paradigm in Fig. \ref{fig:intro} (a) -- neither multiple entrances/exits nor reversible structures are employed. 
Based on those grid-like network topology, NAS based works \citep{ding2021hr, wu2021fbnetv5,liu2019auto,ghiasi2019fpn} search the optimized topology of network architectures for specific dataset. However, the RevCol architecture is not limit to specific tasks or datasets. With the reversible nature, our method maintains lossless information propagation and benefits for not only pre-training but also other down-stream tasks. 
Very recently, \emph{RevBiFPN} \citep{chiley2022revbifpn} comes up with an reversible variant of FPN, which is further employed in an HRNet-like architecture. Though our RevCol shares the similar idea of multi-scale reversible transformations with RevBiFPN, our work is done independently, which is derived from a different motivation of feature disentangling, and has much simpler architectures (\eg free of reversible upsampling tower) and higher performances. We compare some of those models in Sec.~\ref{experiments}.

\section{Method}
\label{method}

In this section, we introduce the design details of our \emph{Reversible Column Networks (RevCol)}. Fig.~\ref{fig:intro} (b) illustrates the top-level architecture. Notice that for each column in RevCol, for simplicity we directly reuse existing structures such as \emph{ConvNeXt} \citep{liu2022convnet}, hence in the following subsections, we mainly focus on how to build the reversible connections between columns. In addition, we introduce an plug-and-play intermediate supervision on top of each column, which further improves the training convergence and feature quality.

\subsection{Multi-Level Reversible Unit}

\label{3:reverse}

\begin{figure}[t]
\begin{center}

\includegraphics[width=1.0\columnwidth]{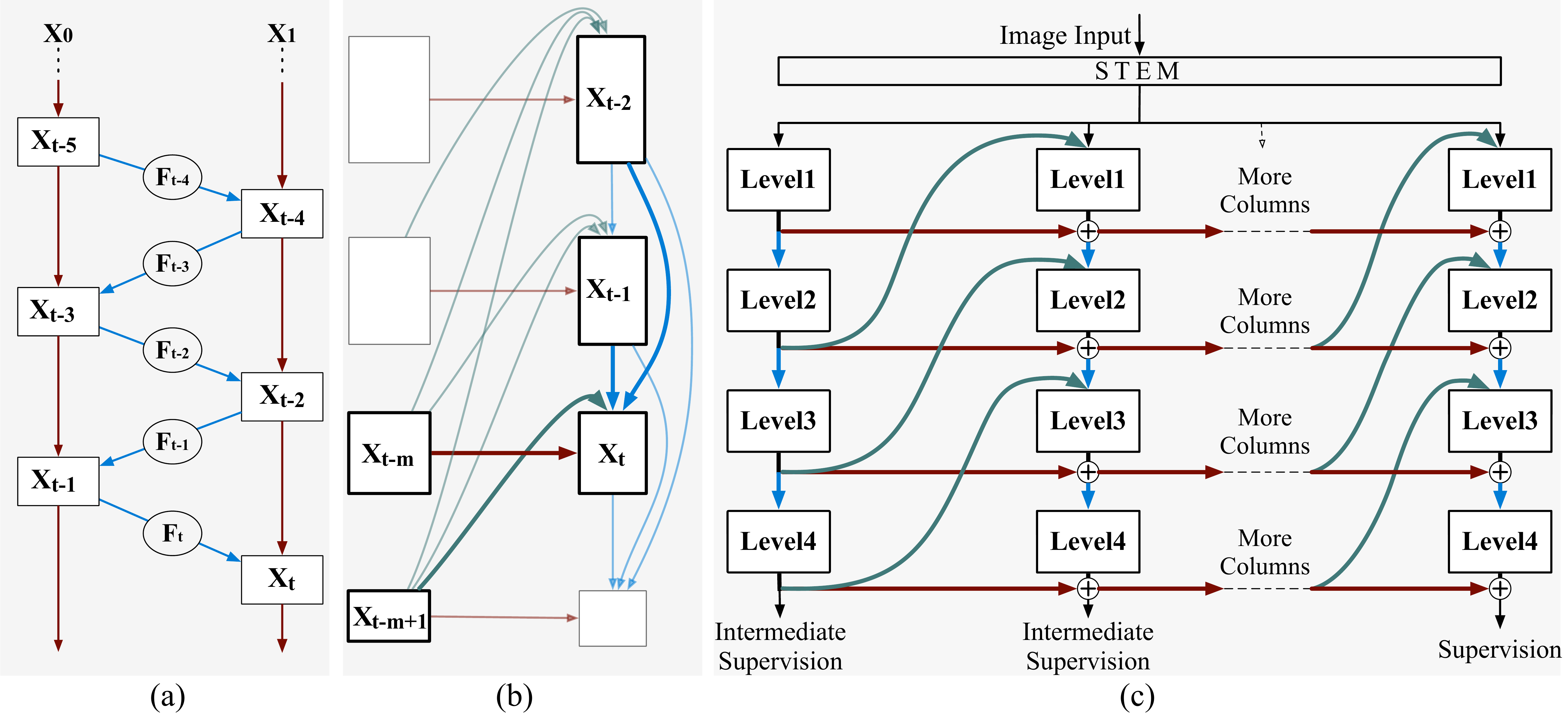}
\end{center}
\caption{(a) Reversible unit in RevNet~\citep{gomez2017reversible}. (b) Multi-level reversible unit. All inputs for level $t$ are highlighted. (c) An overview of the whole reversible column network architecture, with simplified multi-level reversible unit. }
\label{fig:arch}
\end{figure}

In our network, \emph{reversible transformations} plays a key role in feature disentangling without information loss, whose insight comes from \emph{Reversible Neural Networks} \citep{dinh2014nice,chang2018reversible,gomez2017reversible,jacobsen2018revnet,mangalam2022reversible}. 
Among them, we first take a review of one representative work \emph{RevNet} \citep{gomez2017reversible}. As shown in Fig.~\ref{fig:arch} (a), RevNet first partitions the input $x$ into two groups, $x_0$ and $x_1$. Then for later blocks, for example, block $t$, it takes two anterior blocks' outputs $x_{t-1}$ and $x_{t-2}$ as input and generates the output $x_t$. The mapping of block $t$ is \emph{reversible}, i.e. $x_{t-2}$ can be reconstructed by two posterior blocks $x_{t-1}$ and $x_{t}$.
Formally, the forward and \emph{inverse} computation follow the equations \footnote{In \citet{gomez2017reversible}, the proposed reversible equations are formulated as $y_1 = x_1 + \mathcal{F}(x_2)$ and $y_2 = x_2 + \mathcal{G}(y_1)$. While in this paper, we reformulate those notations $y_2, y_1, x_2, x_1, \mathcal{G}, \mathcal{F}$ as $x_t, x_{t-1}, x_{t-2}, x_{t-3}, \boldsymbol{F}_t, \boldsymbol{F}_{t-1}$, respectively, in order to better illustrate the relation between building block $t$ and $t-1$. It is easy to prove the two formulations are equivalent.  }:
\begin{equation}
\begin{aligned} 
    Forward&: x_{t} = \boldsymbol{F}_t(x_{t-1}) + \gamma x_{t-2}\\
    Inverse&: x_{t-2} = \gamma^{-1}[x_{t} - \boldsymbol{F}_t(x_{t-1})] ,
\end{aligned}
\label{eq:revnet}
\end{equation}
where $\boldsymbol{F}_t$ denotes an arbitrary non-linear operation analogous to those residual functions in standard \emph{ResNets}; $\gamma$ is a simple reversible operation (\eg channel-wise scaling), whose inverse is denoted by $\gamma^{-1}$.  
As mentioned in the introduction, the above formulation involves too strong constraint on the feature dimensions, i.e. $x_{t}, x_{t+2}, x_{t+4}, ...$ have to be equal sized, which is not flexible in architecture design. That is why RevNets \citep{gomez2017reversible} introduce some non-reversible down-sampling blocks between reversible units, hence the whole network is not fully reversible. More importantly, we find there is no clear way to directly employ Eq.~\ref{eq:revnet} to bridge the columns in Fig.~\ref{fig:intro} (b).

To address the issue, we generalize Eq.~\ref{eq:revnet} into the following form:
\begin{equation}
\begin{aligned}
Forward&: x_{t} = \boldsymbol{F}_t(x_{t-1}, x_{t-2}, ..., x_{t-m+1}) +  \gamma x_{t-m} \\
Inverse&: x_{t-m} = \gamma^{-1}[x_{t} - \boldsymbol{F}_t(x_{t-1}, x_{t-2}, ..., x_{t-m+1}) ], 
\label{eq:revcol}
\end{aligned}
\end{equation}
where $m$ is the \emph{order} of the recursion $(m\ge 2)$. Clearly, the extension is still reversible. Then we partition every $m$ feature maps into a group: $(x_1, x_2, \dots, x_m), (x_{m+1}, x_{m+2}, \dots, x_{2m}), \dots$. Given the features within any of the group, we can easily compute the features in other groups recursively according to Eq.~\ref{eq:revcol}. Compared with the original form, Eq.~\ref{eq:revcol} has the following two nice properties:
\begin{itemize}
\item The constraint on the feature map sizes is greatly relaxed if $m$ is relatively large. Notice that Eq.~\ref{eq:revnet} does not require feature maps \emph{within} each group to be equal sized; such constraint only exist between groups. Therefore, we can use tensors of different shape to represent features of different semantic levels or different resolutions. 
\item Eq.~\ref{eq:revcol} can easily cooperate with existing network architectures, even though the latter is not reversible. For example, we can assign $m$ feature maps in a standard \emph{ResNet} to represent the feature maps within a group $(x_t, x_{t+1}, \dots, x_{t+m-1})$, which is still compatible with Eq.~\ref{eq:revcol} since ResNet can be viewed as a part of $(\boldsymbol{F}_t, \boldsymbol{F}_{t+1}, \dots, \boldsymbol{F}_{t+m-1})$ respectively. Thus the whole network is still reversible. 
\end{itemize}

Therefore, we can reorganize Eq.~\ref{eq:revcol} into a \emph{multi-column} fashion, as shown in Fig.~\ref{fig:arch} (b). Each column is composed of $m$ feature maps within a group, as well as their mother network. We name it \emph{multi-level reversible unit}, which is the basic component of our RevCol as in Fig.~\ref{fig:intro} (b). 

\subsection{Reversible Column Architecture}
\label{3:arch}

\subsubsection{Macro Design}
\label{sec:macro}

As discussed in the introduction (see Fig.~\ref{fig:intro} (b)), our network \emph{RevCol} is composed of multiple subnetworks with \emph{reversible connections} to perform feature disentangling. Fig.~\ref{fig:arch} (c) elaborates the architecture design. Following the common practice of recent models \citep{dosovitskiy2020image,liu2022convnet}, first the input image is split into non-overlapping patches by a patch embedding module. After that, patches are fed into each subnetwork (column). Columns can be implemented with any conventional single-column architectures, \eg \emph{ViT} \citep{dosovitskiy2020image} or \emph{ConvNeXt} \citep{liu2022convnet}. 
We extract four-level feature maps from each column to propagate information between columns; for example, if the columns are implemented with widely-used hierarchical networks \citep{liu2021swin,he2016deep,liu2022convnet}, we can simply extract multi-resolution features from the output of each stage. 
For classification tasks, we only use feature map of the last level (Level 4) in the last column for rich semantic information. For other down-stream tasks like object detection and semantic segmentation, we use feature maps of all the four levels in the last column as they contain both low-level and semantic information. 

To implement the reversible connections between columns, we adopt the \emph{multi-level reversible unit} proposed in Eq.~\ref{eq:revcol}, but in a simplified fashion: rather than take $(m-1)$ inputs for each non-linear operation $\boldsymbol{F}_t(\cdot)$, we use only one low-level feature $x_{t-1}$ at the current column and one high-level feature $x_{t-m+1}$ at the previous column as the input. The simplification does not break the reversible property. We find more inputs bring minor accuracy gain but consume much more GPU resources. Thus Eq. \ref{eq:revcol} is simplified as:

\begin{equation}
\begin{aligned}
Forward&: x_{t} = \boldsymbol{F}_{t}(x_{t-1}, x_{t-m+1}) +  \gamma x_{t-m} \\
Inverse&: x_{t-m} = \gamma^{-1}[x_{t} - \boldsymbol{F}_{t}(x_{t-1}, x_{t-m+1}) ]. 
\label{eq:simrevcol}
\end{aligned}
\end{equation}

Compared with conventional architectures, the \emph{macro design} of our RevCol has the following three properties or advantages:

\paragraph{Feature disentangling.} In RevCol, the lowest level of each column maintains low-level features as it is close to the input, while the highest level in the last column is highly semantic because it is directly connected to the supervision. Therefore, information in different levels is gradually \emph{disentangled} during the (lossless) propagation between columns -- some feature maps are more and more semantic and some maintain to be low-level. Detailed analyses are presented in Appendix \ref{app:vis}. The property brings many potential advantages, for instance, more flexible to downstream tasks which rely on both high-level and low-level features. We argue that \emph{reversible connection} plays a key role in the disentangling mechanism -- some previous works like \emph{HRNet} \citep{wang2020deep} involve multi-level feature fusion but without reversible connection, which may suffer from information loss and lead to inferior performances in our experiments (see Section~\ref{app:rev_vs_nonrev}). 

\paragraph{Memory saving.} The training of conventional networks takes a lot of memory footprint to store the activations during forward propagation as the demand of gradient computation. While in our RevCol, since the connections between columns are \emph{explicitly} reversible, during the back-propagation we can reconstruct the required activations \emph{on the fly} from the last column to the first, which means we only need to maintain activations from \emph{one} column in memory during training. In Section~\ref{app:gpumemory}, we demonstrate RevCol costs roughly $\mathcal{O}(1)$ additional memory with the increase of column numbers. 

\paragraph{New scaling factor for big models.} In \net~architecture, column number serves as a new dimension in addition to depth (number of blocks), and width (channels of each block) in vanilla single-column CNNs or ViTs. Increasing column numbers has similar income as increasing both width and depth in certain range.

\subsubsection{Micro Design} \label{sec:micro}
We employ \emph{ConvNeXt} blocks \citep{liu2022convnet} to implement each column in our network by default; other architectures, such as \emph{transformers}, are also applicable (see Appendix~\ref{appendix:transformer} for details). We make a few modifications to make ConvNeXt compatible with our macro architecture: 

\paragraph{Fusion module.} As shown in Fig.~\ref{fig:micro}, in each level of original ConvNeXt, the inputs are first down-sampled in a patch-merging block. Then the outputs are passed through a bunch of residual blocks. In RevCol, we introduce a fusion module to fuse the feature maps from the current and previous columns (refer to Fig.~\ref{fig:arch} (c), green and blue connections). We modify the original patch-merging block in ConvNeXt by putting the LayerNorm after the patch-merging convolution rather than before. Channel numbers are doubled in patch-merging convolution. We also introduce an up-sampling block, which is composed of a linear channel mapping layer, a LayerNorm normalization and a feature map interpolation layer. We halve the channel numbers in linear channel mapping layer.
The outputs of the two blocks are summed up and then passed to the residual blocks followed by.

\paragraph{Kernel size.} In RevCol we revise the $7\times 7$ convolutions in original ConvNeXt~\citep{liu2022convnet} to $3\times 3$ by default, mainly to speed up the training. 
Increasing kernel size further obtains more accuracy, but not very much, partly because the our multi-column design enlarges the effective receptive field. Please refer to Section~\ref{sec:ab_ks} for more details. 

\paragraph{Reversible operation $\boldsymbol{\gamma}$.} We adopt a learnable reversible channel-wise scaling as reversible operation $\gamma$ to keep the network stable. Each time the features are summed up in forward of Eq.~\ref{eq:simrevcol}, the magnitude grows larger, which makes the training process unstable. Using a learnable scaling can suppress the magnitude of features. During training, we truncate the absolute value of $\gamma$ so that it will never be smaller than $1e^{-3}$, because the numerical error could become large in the reverse computation when $\gamma$ is too small.

\subsection{Intermediate Supervision}
\label{3:is}
Though \emph{multi-level reversible unit} is able to maintain information during column iteration, the down-sample block still can discard information \emph{inside} column. Features at the end of the front columns is too close to the final output, for reversible connections simply do scaling and summation. Such information loss leads to inferior performance. Similar problem also happens when using deeply-supervised method~\citep{lee2015deeply, szegedy2015going}.

To mitigate the problem of information collapse, we propose an intermediate supervision method which adds additional supervision into front columns.  
For features in front columns, we hope to keep the mutual information between features and the input image as much as possible, so that the network discard less information within columns. Consider \emph{\net} gradually disentangle semantic and low-level information, extracting and leveraging the task-relevant information can further boost the performance. Therefore, we need to maximize the lower bound of mutual information between features and the prediction.

Inspired by \citet{wang2021revisiting}, we add two auxiliary heads to last level features (Level 4). One is a decoder~\citep{he2022masked} which reconstructs the input image, the other is a linear classifier.  
The linear classifier can be trained in a regular classification fashion with the \emph{cross-entropy (CE)} loss.
The parameters of decoder are optimized by minimizing the \emph{binary cross-entropy (BCE)} reconstruction loss.
Compared with commonly used \emph{L1} and \emph{L2} loss, interpreting the distribution of reconstructed logits and input image as \emph{bit probabilities (Bernoullis)} outputs smoother value, which makes it more compatible with CE Loss. 

For intermediate supervision at one column, the compound loss is the weighted summation of the above two loss.
Note that supervision heads may not be added to all columns. For all the variants of \net, we set the number of compound loss to 4 empirically (eg. for a 8 column \net, the supervision heads are added to column 2, 4, and 6, and 8).

The total loss  $\boldsymbol{L}$ in is the summation of all compound loss:
\begin{equation}
\boldsymbol{L} = \sum_{i=1}^{n} ( \alpha_i \mathcal{L}_{\text{BCE}} + \beta_i \mathcal{L}_{\text{CE}} )
\label{eq:is}
\end{equation}
$n$ denotes the total number of compound loss. $\mathcal{L}_{\text{BCE}}$ and $\mathcal{L}_{\text{CE}}$ denotes BCE loss and CE loss correspondingly. $\alpha_i$ and $\beta_i$ are changed linearly with the compound loss number. When the compound loss is added in earlier columns, we use larger value $\alpha_i$ and smaller value $\beta_i$ to keep  $I(\boldsymbol{h},x)$. In later columns, value $\alpha_i$ decreases and $\beta_i$ increases, which helps boost the performance.

\section{Experiments}
\label{experiments}
We construct different \emph{\net}~variants, \net-T/S/B/L, to be of similar complexities to Swin transformers and ConvNeXts. We also build a larger \net-XL and \net-H to test the scaling up capability. These variants adopt different channel dimension $C$, blocks in each column $B$ and column numbers $\textit{COL}$. The configuration hyper-parameters of these model variants are:
\begin{itemize}
 \small
  \item \net-T: \ $C = (64, 128, 256, 512)$, \ $B = (2, 2, 4, 2)$, \ $\textit{COL} = 4$
  \item \net-S: \ $C = (64, 128, 256, 512)$, \ $B = (2, 2, 4, 2)$, \ $\textit{COL} = 8$
  \item \net-B: \ $C = (72, 144, 288, 576)$, \ $B = (1, 1, 3, 2)$, \ $\textit{COL} = 16$
  \item \net-L: \ $C = (128, 256, 512, 1024)$, \ $B = (1, 2, 6, 2)$, \ $\textit{COL} = 8$
  \item \net-XL: \ $C = (224, 448, 896, 1792)$, \ $ B = (1, 2, 6, 2)$, \ $\textit{COL} = 8$
  \item \net-H: \ $C = (360, 720, 1440, 2880)$, \ $ B = (1, 2, 6, 2)$, \ $\textit{COL} = 8$
\end{itemize}

We conduct image classification on \emph{ImageNet} dataset~\citep{deng2009imagenet, ridnik2021imagenet}.
We also test our models on the downstream object detection task and semantic segmentation task on commonly used \emph{MS-COCO} \citep{lin2014microsoft} and \emph{ADE20k} \citep{zhou2017scene} dataset. Training and fine-tuning settings please refer to Appendix~\ref{appendix:details}.  
Furthermore, we show the performance of \net~with transformer on vision and language tasks (shown in Appendix~\ref{appendix:transformer}).

\subsection{Image Classification}
\label{chap:img}

On ImageNet (1.28M images) \citep{deng2009imagenet} dataset, we train \net~ for 300 epochs with intermediate supervision. Hyperparameters, augmentation and regularization strategies follows \citet{liu2022convnet} 
We also pre-train our models on the larger ImageNet-22K dataset \citep{ridnik2021imagenet}, which contains 14.2 million images.

In Tab.~\ref{tab:imagenet}, we compare our \net~ variants with commonly used recent \emph{Transformers} and \emph{CNNs} on ImageNet-1k validation set. 
Our models outperforms a large number of vanilla single-column CNNs and Transformers with similar complexities. For example, \net-S achieve 83.5\% Top-1 accuracy, outperform ConvNeXt-S by 0.4 points. When pre-trained with larger ImageNet-22K dataset, RevCol-XL achieves 88.2\% Top-1 accuracy.  As RevCol maintains some task-irrelevant low-level information in classification pre-training, relaxing the constraint of params and FLOPs and enlarging dataset size can further boost our models' performance. To further test the scaling up effectiveness of large dataset, we build a 168-million-image semi-labeled dataset (see Appendix~\ref{appendix:external-dataset}). With extra data pre-training and ImageNet-1k fine-tuning, our \net-H achieves \textbf{90.0\% top-1 accuracy}. Our results further demonstrate with RevCol, CNN models can also share the dividends of large model and massive data pre-training.

\begin{table}[t]
\centering
\captionsetup{font={small}}
\caption{ \textbf{ImageNet classification results.} We compare our models with state-of-the-art \bv Vision Transformers  and \bc CNNs that have comparable FLOPs and parameters. ↑ denotes models fine-tuning using image size larger than $224^2$. We report the top-1 accuracy on the validation set of ImageNet as well as the number of parameters and FLOPs. Our models are highlighted in \hl{gray}. }

\begin{minipage}{.5\textwidth}
\adjustbox{width=1.0\linewidth}{
\renewcommand\arraystretch{1.1}
\setlength{\tabcolsep}{0pt}
\begin{tabular}{L{122pt}L{35pt}L{30pt}L{30pt}L{30pt}}
\toprule

\multirow{2}[2]{*}{Model} & Image & Params & FLOPs & Top-1 \\ 
& Size & (M) & (G) & Acc. \\
\midrule
\multicolumn{5}{l}{\emph{ImageNet-1K trained models}} \\
\addlinespace[1pt]
\midrule
\bv Swin-T~(\citeauthor{liu2021swin}) & $224^2$ &28 & 4.5 & 81.3 \\
\bv DeiT-S~(\citeauthor{touvron2020training}  & $224^2$ &22 & 4.6 & 79.8 \\
\bv Rev-ViT-S~(\citeauthor{mangalam2022reversible}) & $224^2$ &22 & 4.6 & 79.9 \\
\bc RevBiFPN-S3~~(\citeauthor{chiley2022revbifpn})  & $288^2$ &20 & 3.3 & 81.1 \\
\bc EfficientNet-B4~(\citeauthor{tan2019efficientnet}) & $380^2$ & 19 & 4.2 & 82.9 \\
\bc ConvNeXt-T~(\citeauthor{liu2022convnet}) & $224^2$ &29 & 4.5 & 82.1 \\
\rowcolor{Gray} \bc \net-T & $224^2$ &30 & 4.5 & 82.2 \\

\midrule
\bv Swin-S~(\citeauthor{liu2021swin}) & $224^2$ &50 & 8.7 & 83.0 \\
\bv MViTv1-B~(\citeauthor{fan2021multiscale}) & $224^2$ &37 & 7.8 & 83.0 \\
\bv T2T-ViT-19~(\citeauthor{yuan2021tokens}) & $224^2$ &39 & 8.4 & 81.4\\
\bc RevBiFPN-S4~(\citeauthor{chiley2022revbifpn}) & $320^2$ &49 & 10.6 & 83.0 \\
\bc EfficientNet-B5~ \citeauthor{tan2019efficientnet}) & $456^2$ & 30 & 9.9 & 83.6 \\ 
\bc ConvNeXt-S~(\citeauthor{liu2022convnet}) & $224^2$ &50 & 8.7 & 83.1\\
\rowcolor{Gray} \bc \net-S & $224^2$ &60 & 9.0 & 83.5 \\
\midrule

\bv Swin-B~(\citeauthor{liu2021swin}) & $224^2$ &89 & 15.4 & 83.5 \\
\bv DeiT-B~(\citeauthor{touvron2020training})  & $224^2$ &86 & 17.5 & 81.8 \\
\bv Rev-ViT-B(\citeauthor{mangalam2022reversible}) & $224^2$ &87 & 17.6 & 81.8 \\
\bc RepLKNet-31B~(\citeauthor{ding2022scaling}) & $224^2$ &79 & 15.3 & 83.5 \\
\bc RevBiFPN-S5~(\citeauthor{chiley2022revbifpn}) & $352^2$ &82 & 21.8 & 83.7 \\
\bc EfficientNet-B6~(\citeauthor{tan2019efficientnet}) & $528^2$ & 43 & 19.0 & 84.0 \\ 
\bc ConvNeXt-B~(\citeauthor{liu2022convnet}) & $224^2$ &88 & 15.4 & 83.8 \\
\rowcolor{Gray} \bc \net-B & $224^2$ & 138 & 16.6 & 84.1 \\

\bottomrule

\end{tabular}%
}
\end{minipage}
\hfill
\begin{minipage}{.49\textwidth}
\adjustbox{width=1.0\linewidth}{
\renewcommand\arraystretch{1.1}
\setlength{\tabcolsep}{0pt}
\begin{tabular}{L{125pt}L{35pt}L{30pt}L{30pt}L{30pt}}\toprule

\multirow{2}[2]{*}{Model} & Image & Params & FLOPs & Top-1 \\ 
& Size & (M) & (G) & Acc. \\
\midrule
\multicolumn{5}{l}{\emph{ImageNet-22K pre-trained models (ImageNet-1K fine-tuned) }} \\
\addlinespace[1pt]
\midrule
\bv Swin-B~(\citeauthor{liu2021swin} & $224^2$ & 88 & 15.4 & 85.2 \\
\bv Swin-B↑~(\citeauthor{liu2021swin} & $384^2$ & 88 & 47.0 & 86.4 \\
\bv ViT-B↑~(\citeauthor{dosovitskiy2020image}) & $384^2$ & 86 & 55.4 & 84.0 \\
\bc RepLKNet-31B~(\citeauthor{ding2022scaling}) & $224^2$ & 79 & 15.3 & 85.2 \\ 
\bc RepLKNet-31B↑~(\citeauthor{ding2022scaling}) & $384^2$ & 79 & 45.1 & 86.0 \\ 
\bc ConvNeXt-B~(\citeauthor{liu2022convnet}) & $224^2$ & 89 & 15.4 & 85.8\\
\bc ConvNeXt-B↑~(\citeauthor{liu2022convnet}) & $384^2$ & 89 & 45.1 & 86.8\\
\rowcolor{Gray} \bc \net-B & $224^2$ & 138 & 16.6 & 85.6 \\
\rowcolor{Gray} \bc \net-B↑ & $384^2$ & 138 & 48.9 & 86.7 \\


\midrule
\bv Swin-L~(\citeauthor{liu2021swin}) & $224^2$ & 197 & 34.5 & 86.3 \\
\bv Swin-L↑~(\citeauthor{liu2021swin}) & $384^2$ & 197 & 103.9 & 87.3 \\
\bv ViT-L↑~(\citeauthor{dosovitskiy2020image}) & $384^2$ & 307 & 190.7 & 85.2 \\

\bc RepLKNet-31L~(\citeauthor{ding2022scaling}) & $384^2$ & 172 & 96.0 & 86.6 \\ 
\bc ConvNeXt-L~(\citeauthor{liu2022convnet}) & $224^2$ & 198 & 34.4 & 86.6 \\
\bc ConvNeXt-L↑~(\citeauthor{liu2022convnet}) & $384^2$ & 198 & 101.0 & 87.5 \\
\rowcolor{Gray} \bc \net-L & $224^2$ & 273 & 39.0 & 86.6 \\
\rowcolor{Gray} \bc \net-L↑ & $384^2$ & 273 & 116.0 & 87.6 \\
\midrule
\bc ConvNeXt-XL↑ ~(\citeauthor{liu2022convnet}) & $384^2$ & 350 & 179.0 &87.8  \\
\rowcolor{Gray} \bc \net-XL↑ & $384^2$ & 834 & 350.0 &  88.2  \\
\midrule
\multicolumn{5}{l}{\emph{Extra data pre-trained models (ImageNet-1K fine-tuned) }} \\
\addlinespace[1pt]
\midrule 
\rowcolor{Gray} \bc \net-XL↑ & $384^2$ & 834 & 350.0 & 89.4 \\
\rowcolor{Gray} \bc \net-H↑ & $640^2$ & 2158 & 2537 & 90.0 \\
\bottomrule

\end{tabular}%
}
\end{minipage}

\label{tab:imagenet}%
  \vspace{-10pt}
\end{table}%

\subsection{Object Detection}
\label{chap:coco}

We evaluate our proposed \net~on object detection task. Experiments are conducted on the MS-COCO dataset using the Cascade Mask R-CNN \citep{cai2019cascade} framework. We also finetune our largest model RevCol-H with HTC++~\citep{chen2019hybrid} and DINO~\citep{zhang2022dino} Framework.
\begin{table}[t]
 \centering 
 \captionsetup{font={small}}
 \caption{ \textbf{Object detection results on MS-COCO dataset with different backbones.} We report box AP and mask AP with single scale testing on COCO minival set. FLOPs are measured under input sizes of (1280, 800). }
 \adjustbox{width=0.94\linewidth}{
  \begin{tabular}{llccccclr}
  \toprule 
    Backbone & AP$^{ box}$ & AP$^{ box}_{50}$ & AP$^{ box}_{75}$ & AP$^{ mask}$ & AP$^{ mask}_{50}$ &  AP$^{ mask}_{75}$ & Params & FLOPs \\
\midrule
\multicolumn{9}{c}{\emph{ImageNet-1K pre-trained}} \\
\hline
\addlinespace[2pt]
\bv Swin-T~(\citeauthor{liu2021swin}) & 50.5 & 69.3 & 54.9 & 43.7 & 66.6 & 47.1 & 86M  & 745G  \\
\bc ConvNeXt-T~(\citeauthor{liu2022convnet}) & 50.4 & 69.1 & 54.8 & 43.7 & 66.5 & 47.3 & 86M & 741G  \\
\rowcolor{Gray} \bc  \net-T          & 50.6 & 68.9 & 54.9 & 43.8 & 66.7 & 47.4 & 88M &741G   \\
\bv Swin-S~(\citeauthor{liu2021swin}) & 51.8 & 70.4 & 56.3 & 44.7 & 67.9 & 48.5 & 107M & 838G \\
\bc ConvNeXt-S~(\citeauthor{liu2022convnet}) & 51.9 & 70.8 & 56.5 & 45.0 & 68.4 & 49.1 & 108M & 827G \\
\rowcolor{Gray} \bc \net-S           & 52.6 & 71.1 & 56.8 & 45.5 & 68.8 & 49.0 & 118M & 833G  \\
\bv Swin-B~(\citeauthor{liu2021swin}) & 51.9 & 70.9 & 56.5 & 45.0 & 68.4 & 48.7 & 145M & 982G \\
\bc ConvNeXt-B~(\citeauthor{liu2022convnet})  & 52.7 & 71.3 & 57.2 & 45.6 & 68.9 & 49.5 & 146M & 964G \\
\bc RepLKNet-B~(\citeauthor{ding2022replknet}) &  52.2 & - & - & 45.2  & - & - & 137M & 965G  \\
\rowcolor{Gray} \bc \net-B            & 53.0 & 71.4 & 57.3 & 45.9 & 69.1 & 50.1 & 196M & 988G   \\

\midrule
\multicolumn{9}{c}{\emph{ImageNet-22K pre-trained}} \\
\hline
\addlinespace[2pt]
\bv Swin-B ~(\citeauthor{liu2021swin})& 53.0 & 71.8 & 57.5 & 45.8 & 69.4 & 49.7 & 145M & 982G \\
\bc ConvNeXt-B~(\citeauthor{liu2022convnet})  & 54.0 & 73.1 & 58.8 & 46.9 & 70.6 & 51.3 & 146M & 964G \\
\bc RepLKNet-B~(\citeauthor{ding2022replknet}) &  53.0 & - & - & 46.3  & - & - & 137M & 965G  \\

\rowcolor{Gray} \bc \net-B            & 55.0 & 73.5 & 59.7 & 47.5 & 71.1 & 51.8 & 196M & 988G  \\
\bv Swin-L ~(\citeauthor{liu2021swin}) &53.9 & 72.4 & 58.8 & 46.7 & 70.1 & 50.8 & 253M & 1382G \\
\bc ConvNeXt-L ~(\citeauthor{liu2022convnet})  &54.8 & 73.8 & 59.8 & 47.6 & 71.3 & 51.7 & 255M & 1354G \\
\bc RepLKNet-L~(\citeauthor{ding2022replknet}) & 53.9  & - & - & 46.5  & - & - & 229M & 1321G \\
\rowcolor{Gray} \bc \net-L             &55.9 & 74.1 & 60.7 & 48.4 & 71.8 & 52.8 & 330M & 1453G  \\
  
  \midrule
\multicolumn{9}{c}{\emph{Extra data pre-trained}} \\
\hline
\addlinespace[2pt]
\rowcolor{Gray} \bc \net-H (HTC++) &61.1&	78.8&	67.0 & 53.0 & 76.3 & 58.7 & 2.41G & 4417G  \\

\rowcolor{Gray} \bc \net-H (Objects365+DINO) &63.8&	81.8&	70.2 & - & - & - & 2.18G & 4012G  \\
  \bottomrule

  \end{tabular}%
  }
  \vspace{-8pt}

 \label{tab:downstream}%
\end{table}%

In Tab. \ref{tab:downstream}, we compare the \emph{AP$_{box}$} and \emph{AP$_{mask}$} with Swin/ConvNeXt in variant sizes on COCO validation set. 
We find RevCol models surpass other counterparts with similar computation complexities. Information retained in pre-training helps RevCol models acheieve better results in down-stream tasks. When the model size grows larger, this advantage becomes more remarkable. After finetuning under Objects365\citep{shao2019objects365} dataset and DINO~\citep{zhang2022dino} framework, our largest model \net-H achieves \textbf{63.8\%} AP$_{box}$ on COCO detection minival set.
\begin{table}[ht]
 \centering  
 \captionsetup{font={small}}
 \caption{ \textbf{Semantic segmentation result on ADE20k dataset with different backbones.} 
 we report mIoU results with single/multi-scale testing. FLOPs are measured under input sizes of (2048, 512), (2560, 640) for IN-1K and IN-22K pre-trained models respectively.}
 
 \adjustbox{width=0.77\linewidth}{
  \begin{tabular}{lllllc}
  \toprule 
    Backbone & crop size & $mIoU_{ss}$ & $mIoU_{ms}$ & Params & FLOPs \\
\midrule
\multicolumn{6}{c}{\emph{ImageNet-1K pre-trained}} \\
\hline
\addlinespace[2pt]
\bv Swin-T~(\citeauthor{liu2021swin}) & $512^2$ &44.5 & 45.8 & 60M & 945G \\
\bc ConvNeXt-T~(\citeauthor{liu2022convnet})  & $512^2$& 46.0 & 46.7 & 60M & 939G  \\
\rowcolor{Gray} \bc  \net-T          &  $512^2$& 47.4 & 47.6 & 60M & 937G \\

\bv Swin-S~(\citeauthor{liu2021swin}) & $512^2$ & 47.6 & 49.5 & 81M & 1038G \\
\bc ConvNeXt-S~(\citeauthor{liu2022convnet}) & $512^2$ & 48.7 & 49.6 & 82M & 1027G \\
\rowcolor{Gray} \bc \net-S           & $512^2$ & 47.9 & 49.0 & 90M & 1031G \\

\bv Swin-B~(\citeauthor{liu2021swin}) & $512^2$ & 48.1 & 49.7 & 121M & 1188G \\
\bc RepLKNet-B~(\citeauthor{ding2022replknet}) & $512^2$  & 49.9  & 50.6  & 112M & 1170G \\
\bc ConvNeXt-B~(\citeauthor{liu2022convnet})  & $512^2$ & 49.1 & 49.9 & 122M & 1170G \\

\rowcolor{Gray} \bc \net-B            & $512^2$ & 49.0 & 50.1 & 122M & 1169G \\

\midrule
\multicolumn{6}{c}{\emph{ImageNet-22K pre-trained}} \\
\hline
\addlinespace[2pt]
\bv Swin-B ~(\citeauthor{liu2021swin})& $640^2$ & 50.3 & 51.7 & 121M & 1841G \\
\bc RepLKNet-B~(\citeauthor{ding2022replknet})& $640^2$  & 51.5  & 52.3  & 112M & 1829G \\
\bc ConvNeXt-B~(\citeauthor{liu2022convnet})  & $640^2$ & 52.6  & 53.1 & 122M & 1828G \\
\rowcolor{Gray} \bc \net-B            & $640^2$ & 52.7  & 53.3 & 122M & 1827G \\

\bv Swin-L ~(\citeauthor{liu2021swin}) & $640^2$ &  52.1 & 53.5 & 234M & 2468G \\
\bc RepLKNet-L~(\citeauthor{ding2022replknet}) & $640^2$  & 52.4  & 52.7  & 207M & 2404G \\
\bc ConvNeXt-L ~(\citeauthor{liu2022convnet})  & $640^2$ &  53.2 & 53.7 & 235M & 2458G \\
\rowcolor{Gray} \bc \net-L             & $640^2$ &  53.4 & 53.7 & 306M & 2610G \\

\midrule
\multicolumn{6}{c}{\emph{Extra data pre-trained}} \\
\hline
\addlinespace[2pt]
\rowcolor{Gray} \bc \net-H & $640^2$  & 57.8  & 58.0  & 2421M & -\\
\rowcolor{Gray} \bc \net-H + Mask2Former & $640^2$  & 60.4  & 61.0  & 2439M & -\\
  \bottomrule

  \end{tabular}%
  }
\vspace{-5pt}
 \label{tab:ade}%
\end{table}%

\subsection{Semantic Segmentation}

We also evaluate \net~backbones on the ADE20K semantic segmentation task with \emph{UperNet}~\citep{xiao2018unified} framework. We do not use intermediate-supervision in down-stream fine-tune process. To further explore our model's capacity and reach the leading performance, we utilize recent segmentation framework \emph{Mask2Former}~\citep{cheng2022masked}, and adopt the same training settings. 

In Tab. \ref{tab:ade}, we report validation \emph{mIoU} with single-scale and multi-scale flip testing. \net~models can achieve competitive performance across different model capacities, further validating the effectiveness of our architecture design.  It's worth noting that when use Mask2Former detector and extra pre-training data, \net-H achieves an mIoU of 61.0\%, which shows feasible scalability towards large-scale vision applications.
\begin{table}[t]
\centering
\captionsetup{font={small}}
\caption{ \textbf{System-level comparison of state-of-the-art visual foundation models with large-scale pretraining.} We include \bv Vision Transformers, \bc CNNs, and \bh hybrid architectures pretrained either \emph{unsupervised} or \emph{supervised} on \emph{image-only} and \emph{vision-language} datasets. COCO scores marked with $\dagger$ means intermediate fine-tuned on extra data like Object365~\citep{shao2019objects365}.}
\small

\renewcommand\arraystretch{1.0}
\setlength{\tabcolsep}{2pt}
\scalebox{0.88}{
\begin{tabular}
{L{47pt}C{30pt}R{35pt}C{75pt}C{38pt}C{35pt}C{25pt}C{30pt}C{50pt}C{25pt}C{25pt}}

\toprule
\multirow{2}[2]{*}{\;\;\,Model} & \multirow{2}[2]{*}{Params} & \multicolumn{2}{c}{Dataset}  & \multicolumn{1}{c}{ImageNet} & \multicolumn{3}{c}{COCO test-dev} & \multicolumn{3}{c}{ADE20K} \\
 \cmidrule(lr){3-4} \cmidrule(lr){5-5}  \cmidrule(lr){6-8} \cmidrule(lr){9-11}
               &     & Images & Annotation  &  1k & Detector &AP$_{box}$ & AP$_{mask}$ & Segmenter & mIoU & +ms  \\ 
\midrule
\bv SwinV2-G                  & 3.0 G    & 70 M          & labeled                     & 90.2   &  HTC++ & 63.1\rlap{$^\dagger$}             & 54.4\rlap{$^\dagger$}       &UperNet    & 59.3         & 59.9        \\
\bv BEiT3                      & 1.0 G   & 35 M          & labeled \& image-text       & 89.6    &  ViTDet  & 63.7\rlap{$^\dagger$}            & 54.8\rlap{$^\dagger$}        &Mask2Former    & 62.0         & 62.8        \\
\bh Florence         & 0.9 G   & 900 M         & image-text                        & 90.1    &  DyHead &        62.4         &   -         &   -             &      -        &    -         \\
\rowcolor{Gray} \bc RevCol-H                   & 2.1 G    & 168 M         & semi-labeled   & 90.0  &  DINO   &     63.6\rlap{$^\dagger$}            &       -     &Mask2Former     &       60.4       &     61.0        \\

\bottomrule
\end{tabular}
}
\label{tab:sota1}
\end{table}

\subsection{System-level Comparison with SOTA Foundation Models}

Foundation models~\citep{kolesnikov2020big, radford2021learningclip, yuan2021florence} are general-purpose backbones pre-trained on massive and divergent data source. They can adapt to various down-stream tasks with limited domain-specific data. We show comparison among various public \emph{state-of-the-art (SOTA)} foundation models including Vision Transformers and Vision-Language models, namely, \emph{SwinV2}~\citep{liu2022swin}, \emph{BEiT3}~\citep{wang2022beit3}, and \emph{Florence}~\citep{yuan2021florence}. As shown in Tab.~\ref{tab:sota1}, though our \net-H is \emph{purely convolutional and pre-trained on single modality dataset}, the results on different tasks demonstrate remarkable generalization ability of RevCol with large scale parameters.

\subsection{More Analysis Experiments} \label{appendix:experiments}

\subsubsection{Performance Gain of Reversible Columns Architecture }

In this section, we evaluate the performance gain of using reversible columns. In the first experiment, we fix a single column's structure and FLOPs then simply add more columns to scale large and test the performance. At the same time, we plot the vanilla single-column models with similar model sizes. As depicted in Fig.~\ref{fig:multi_vs_single}, compared to single-column models, using multi-column reversible architecture always gets better performance under same FLOPs constraint. Besides, within a certain range, scaling up \net~in terms of increasing column numbers can have similar gains compared to scaling up with both block numbers(depth) and channel numbers(width) in single-column models.  
In the second experiment, we limit the  model size to about 4.5G FLOPs and test model variants with different column numbers. In other words, we gradually add more columns and scale down the single column size at the same time. Results are shown in Tab.~\ref{tab:abcol}, we notice that adopt column number at the range of 4 to 12 can maintain the model's performance, then further more column models suffer from performance degradation. We believe the reason is the width and depth in a single column are too low to keep representation ability.
\begin{table}[h]

 \centering 
 \captionsetup{font={small}}
 \begin{minipage}{.48\textwidth}
\includegraphics[width=0.95\columnwidth]{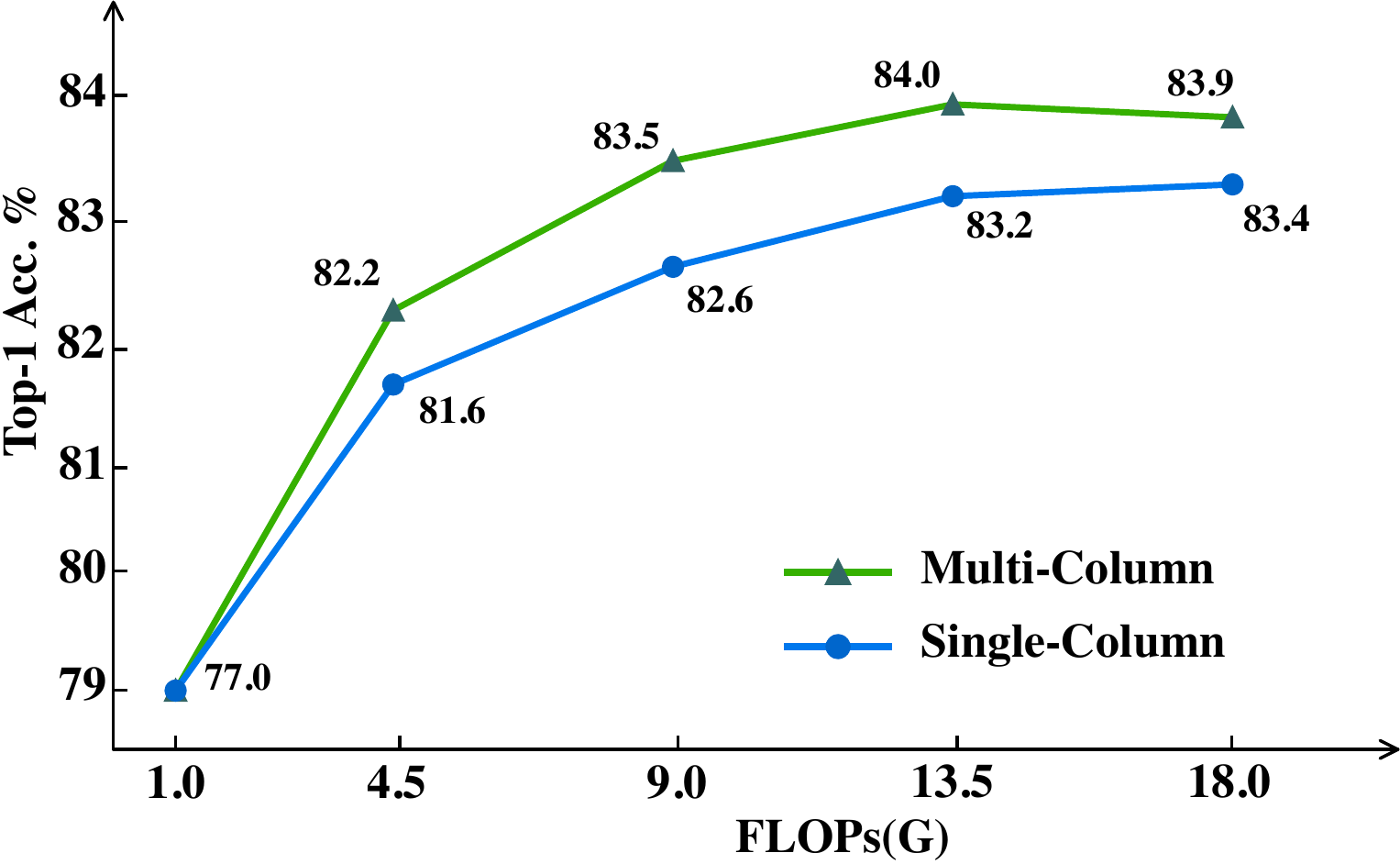}
\vspace{-2mm}
 \captionof{figure}{ImageNet-1K performance of maintaining a constant FLOPs of a single column and adding more columns.}
\label{fig:multi_vs_single}
\end{minipage}
\hfill
 \begin{minipage}{.45\textwidth}
\caption{ImageNet 1K performances of various number of columns in \net{s} under the similar computational budget.}
 \adjustbox{width=\linewidth}{
 \setlength{\tabcolsep}{6pt}
\renewcommand\arraystretch{1.1}

\begin{tabular}{ccccc}
\toprule 
\# column & Params & FLOPs & \begin{tabular}[c]{@{}l@{}}FLOPs\\ per col.\end{tabular} & \begin{tabular}[c]{@{}l@{}}Top-1\\ Acc.\end{tabular} \\ 
\midrule
1      & 28M    & 4.4G  & 4.40G                                                    & 81.9 \\ 
4      & 30M    & 4.5G  & 1.12G                                                   & 82.2 \\ 
8      & 34M    & 4.7G  & 0.59G                                                   & 82.3 \\ 
12     & 33M    & 4.4G  & 0.35G                                                   & 82.2 \\ 
20     & 35M    & 4.2G & 0.21G                                                   & 81.0 \\ 
\bottomrule
\end{tabular}
\hfill
}
 \label{tab:abcol}%
\end{minipage}
\end{table}

\subsubsection{Reversible Networks vs. Non-Reversible Networks}
\label{app:rev_vs_nonrev}

In this section, we ablate different design patterns of reversible connections. First, we build a non-reversible multi-column network using the fusion module of HRNet. Second, we build another single column reversible ConvNeXt using the design of RevNet as shown in Fig. \ref{fig:arch}(a).  We compare the two designs with our \net{s}. The evaluation result is shown in Tab.~\ref{tab:rev}. 
The non-reversible multi-column network suffers from information loss during propagation, which could result in lower accuracy. The reversible single-column network maintains information during propagation, but lack the superiority of multi-level fusion. This experiment further indicates the effectiveness of combining the reversible design with multi-column networks.

\begin{table}[h]
\captionsetup{font={small}}

\begin{minipage}{0.4\linewidth}
\caption{Performance comparison on ImageNet-1K of different design patterns. Row-1 represents HRNet style network w/o \emph{reversible connections}. Row-2 represents RevNet style network w/o \emph{multi-column fusions}. Row-3 are our proposed \net{s}.}
\vspace{7pt}
\renewcommand\arraystretch{1.1}
\setlength{\tabcolsep}{2pt}
\centering
\small
\begin{tabular}{ccccc}
\toprule
 \emph{rev. conn.} & \emph{multi-col.}  & Params & FLOPs & Acc. \\
\midrule
& \vmark    & 35M    & 4.9G  & 78.8       \\ 
\vmark &     & 34M    & 4.5G  & 81.6      \\ 
\vmark & \vmark        & 30M    & 4.5G  & 82.2       \\
\bottomrule
\end{tabular}
\label{tab:rev}
\end{minipage}
\hfill
\begin{minipage}{0.57\linewidth}
\caption{Performance comparison between models with and without \emph{intermediate supervision}. Results are reported on ImageNet-1K and COCO dataset. We use $1\!\times$ training schedule on COCO detection task.}

\renewcommand\arraystretch{1.0}
\setlength{\tabcolsep}{3pt}
\centering
\small
\begin{tabular}{lclll}
\toprule

Model & \emph{inter. sup.} & Top-1 Acc. &  AP$_{box}$ & AP$_{mask}$ \\
\hline 
\noalign{\vskip 0.1cm}
\net-T     & \xmark       & 81.4         & 48.3                & 41.8     \\
\net-T     & \vmark       & 82.2 (+0.8)   & 48.8 (+0.6)          & 42.2 (+0.4)           \\
\hline 
\noalign{\vskip 0.1cm}
\net-S     & \xmark       & 83.0         & 50.7                & 43.8           \\
\net-S     & \vmark       & 83.5 (+0.5)  & 51.1 (+0.4)          & 43.8 (+0.0)          \\
\hline 
\noalign{\vskip 0.1cm}
\net-B     & \xmark       & 83.2         & 51.2                & 44.2          \\
\net-B     & \vmark       & 84.1 (+0.9)   & 51.6 (+0.4)          & 44.2 (+0.0)          \\
\bottomrule
\end{tabular}
\label{tab:intersup}
\end{minipage}
\end{table}

\subsubsection{Performance gain of using Intermediate Supervision}
In this section, we evaluate the performance of \net-T/S/B with and without intermediate supervision on ImageNet-1K. We also evaluate the object detection task performance using $1\times$ training schedule on MS-COCO dataset. 
Other settings remain the same.
From the validation results in Tab. \ref{tab:intersup}, models trained with intermediate supervision achieves 0.5\% to 0.9\% better top-1 accuracy. Besides, intermediate supervision also benefits down-stream tasks, which further demonstrates its effectiveness.

\begin{table}[h]

 \centering 
 \captionsetup{font={small}}
 \begin{minipage}{.55\textwidth}
\includegraphics[width=0.85\columnwidth]{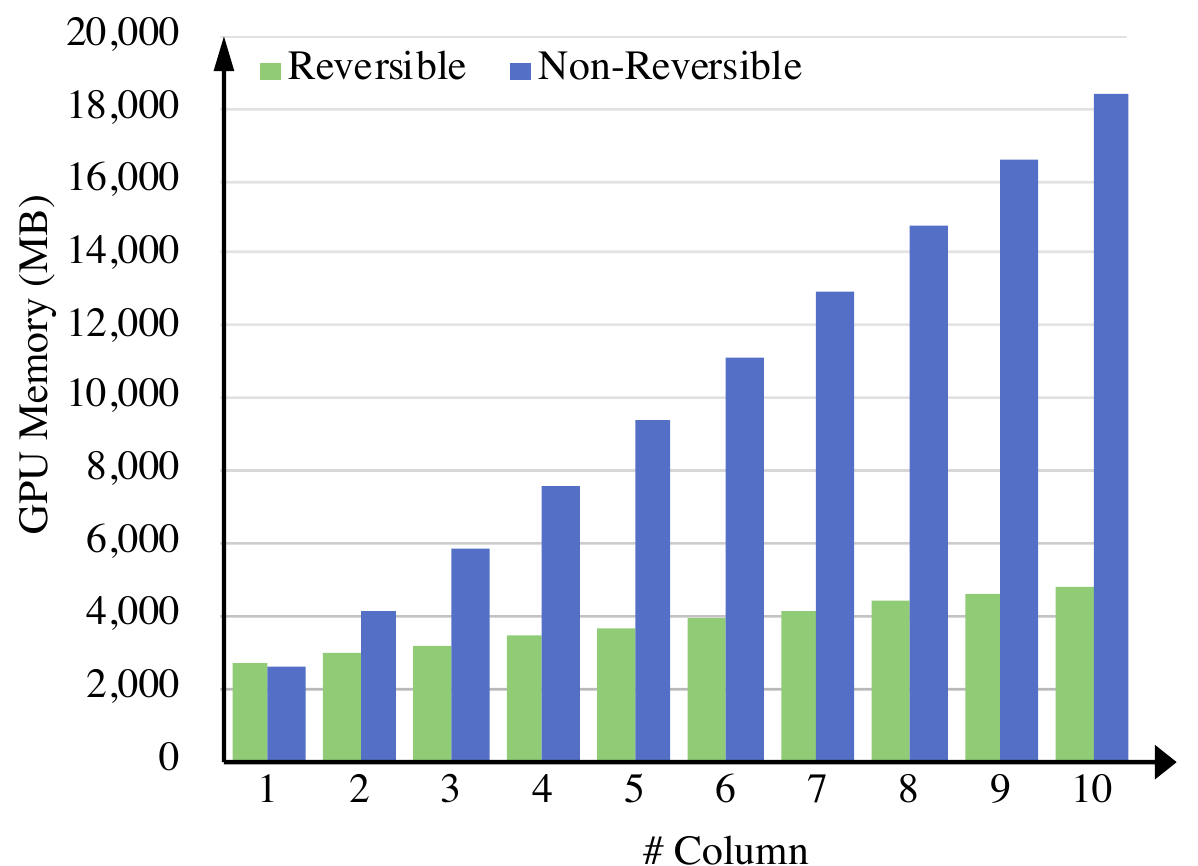}

 \captionof{figure}{GPU Memory Consumption \vs Model size}
\label{fig:mem}

\end{minipage}
\hfill
 \begin{minipage}{.44\textwidth}
\caption{Performance of models with larger kernel convolution.}

 \adjustbox{width=\linewidth}{
 \setlength{\tabcolsep}{4pt}
\renewcommand\arraystretch{1.1}

\begin{tabular}{lcccc}
\toprule
\begin{tabular}[c]{@{}l@{}}Kernel \\ Size\end{tabular} & FLOPs & 
\multicolumn{1}{c}{\begin{tabular}[c]{@{}c@{}}Top-1\\ Acc\end{tabular}} & \multicolumn{1}{c}{\begin{tabular}[c]{@{}c@{}}AP$_{box}$\\ $1\!\times$ \end{tabular}} & \multicolumn{1}{c}{\begin{tabular}[c]{@{}c@{}}AP$_{mask}$\\ $1\!\times$ \end{tabular}} \\ 
\midrule
3  & 4.5G  & 82.2 & 48.8    & 42.2     \\ \midrule
5  & 4.5G  & 82.5 & 49.5    & 42.6     \\ \midrule
7  & 4.6G  & 82.5  & 49.3    & 42.4     \\ \midrule
11 & 4.6G  & 82.5  & 49.9    & 42.7    \\
\bottomrule
\end{tabular}
\hfill
}
 \label{tab:kernel}%
 
\end{minipage}
\end{table}

\vspace{-3pt}
\subsubsection{GPU Memory Consumption vs Model size}
\vspace{-3pt}
\label{app:gpumemory}

Fig. \ref{fig:mem} plots the GPU memory consumption with the scaling of model size. We fix the computation complexity of a single column to 1G FLOPs and increase column number. Meanwhile, we measure the memory consumption in training process which includes the forward and backward propagation. Our experiments are conducted on Nvidia Tesla V100 GPU under batch-size 64, FP16 precision and PyTorch implementation. With the increment of column number, we can see \net~keeps an $\mathcal{O}(1)$ GPU memory consumption, while non-reversible architecture's memory consumption increase linearly with column number. Note that our \net~does not keep strictly the same GPU memory consumption as column number increase, as reversible networks need to back-up the operation weights in need for calculating gradients and the re-construction of feature maps in backward propagation. 

\vspace{-3pt}
\subsubsection{Ablation of kernel size in convolutions} \label{sec:ab_ks}
\vspace{-3pt}
In original ConvNeXt, large kernel convolution achieves in better performance. We conduct experiments in \net-T. As shown in Tab. \ref{tab:kernel}, for 4 column models, using $5\times 5$ convolution increase the ImageNet-1k Top-1 accuracy by 0.3\% and the COCO $AP_{box}$ by 0.7 for \net-T model. Further increasing kernel size obtains more accuracy in down-stream tasks, but not too much.
We consider the RevCol design already enlarges the effective receptive field and this limit the accuracy gain of using large kernel convolution. 
On the other hand, $3\times 3$ convolution enjoys the merits of efficiency and stability in (pre)training. Therefore, we adopt kernel 3 in all RevCol models.

\section{Related Works}
\label{related_works}

\subsection{Disentangle Representation Learning and Part-Whole Hierarchy}
A disentangled representation is generally described as one which separates the factors of variation, explicitly representing the important attributes of the data~\citep{desjardins2012disentangling, bengio2013representation}. 
\cite{desjardins2012disentangling, kulkarni2015deep, higgins2017betavae, chen2016infogan, karras2019style} seek to learn disentangled representations through generative models. \cite{locatello2019challenging} points out that unsupervised learning of disentangled representations is fundamentally impossible without inductive biases both on the considered learning approaches and the datasets. 
The recent proposal of \emph{GLOM}~\citep{hinton2021represent} gives an idea of representing a part-whole hierarchy by a weight-sharing columns. 
The GLOM architecture provides an interpretable part-whole hierarchies for deep neural network~\citep{garau2022interpretable}.
In \net, we adopt the design of using columns, but not modeling the process of formulating islands. 
On the contrary, our column iteration process maintains both low-level and high-level information and gradually disentangle them. Rather than using self-supervised methods, \net~can be trained with supervision end-to-end.

\subsection{Reversible Networks}
\cite{gomez2017reversible} firstly propose \emph{RevNet} that allow back propagation without saving intermediate activations. The reversible design remarkably saves the training cost, since it keep $\mathcal{O}(1)$ GPU memory consumption as model depth scaling up. \cite{jacobsen2018revnet} propose a fully reversible network that can reverse back to the input without any information loss. \cite{chang2018reversible} develop a theoretical framework on stability and reversibility of deep neural network and derive reversible networks that can go arbitrarily deep. \cite{mangalam2022reversible} expand the reversible network scope from CNNs to Transformers.
\emph{RevBiFPN} \citep{chiley2022revbifpn}, a concurrent work of ours, add the reversible connections to \emph{BiFPN}~\citep{tan2020efficientdet} network. Our \net~maintains the information without loss inside each column rather than the whole BiFPN network in RevBiFPN. 
\vspace{-3pt}

\section{Conclusion}
In this paper, we propose \net, a reversible column based foundation model design paradigm. During the lossless propagation through columns, features in RevCol are learned to be gradually disentangled and the total information is still maintained rather than compressed. Our experiments suggests that RevCol can achieve competitive performance in multiple computer vision tasks. We hope RevCol could contribute to better performance in various tasks in both vision and language domains.

\clearpage
\bibliography{reference}

\begin{thebibliography}{78}
\providecommand{\natexlab}[1]{#1}
\providecommand{\url}[1]{\texttt{#1}}
\expandafter\ifx\csname urlstyle\endcsname\relax
  \providecommand{\doi}[1]{doi: #1}\else
  \providecommand{\doi}{doi: \begingroup \urlstyle{rm}\Url}\fi

\bibitem[Bai et~al.(2020)Bai, Koltun, and Kolter]{bai2020multiscale}
Shaojie Bai, Vladlen Koltun, and J~Zico Kolter.
\newblock Multiscale deep equilibrium models.
\newblock \emph{Advances in Neural Information Processing Systems},
  33:\penalty0 5238--5250, 2020.

\bibitem[Bao et~al.(2021)Bao, Dong, and Wei]{bao2021beit}
Hangbo Bao, Li~Dong, and Furu Wei.
\newblock Beit: Bert pre-training of image transformers.
\newblock \emph{arXiv preprint arXiv:2106.08254}, 2021.

\bibitem[Bengio et~al.(2013)Bengio, Courville, and
  Vincent]{bengio2013representation}
Yoshua Bengio, Aaron Courville, and Pascal Vincent.
\newblock Representation learning: A review and new perspectives.
\newblock \emph{IEEE transactions on pattern analysis and machine
  intelligence}, 35\penalty0 (8):\penalty0 1798--1828, 2013.

\bibitem[Cai \& Vasconcelos(2019)Cai and Vasconcelos]{cai2019cascade}
Zhaowei Cai and Nuno Vasconcelos.
\newblock Cascade r-cnn: high quality object detection and instance
  segmentation.
\newblock \emph{IEEE transactions on pattern analysis and machine
  intelligence}, 43\penalty0 (5):\penalty0 1483--1498, 2019.

\bibitem[Caruana(1997)]{caruana1997multitask}
Rich Caruana.
\newblock Multitask learning.
\newblock \emph{Machine learning}, 28\penalty0 (1):\penalty0 41--75, 1997.

\bibitem[Chang et~al.(2018)Chang, Meng, Haber, Ruthotto, Begert, and
  Holtham]{chang2018reversible}
Bo~Chang, Lili Meng, Eldad Haber, Lars Ruthotto, David Begert, and Elliot
  Holtham.
\newblock Reversible architectures for arbitrarily deep residual neural
  networks.
\newblock In \emph{Proceedings of the AAAI conference on artificial
  intelligence}, volume~32, 2018.

\bibitem[Chen et~al.(2019)Chen, Pang, Wang, Xiong, Li, Sun, Feng, Liu, Shi,
  Ouyang, et~al.]{chen2019hybrid}
Kai Chen, Jiangmiao Pang, Jiaqi Wang, Yu~Xiong, Xiaoxiao Li, Shuyang Sun,
  Wansen Feng, Ziwei Liu, Jianping Shi, Wanli Ouyang, et~al.
\newblock Hybrid task cascade for instance segmentation.
\newblock In \emph{Proceedings of the IEEE/CVF Conference on Computer Vision
  and Pattern Recognition}, pp.\  4974--4983, 2019.

\bibitem[Chen et~al.(2015)Chen, Goodfellow, and Shlens]{chen2015net2net}
Tianqi Chen, Ian Goodfellow, and Jonathon Shlens.
\newblock Net2net: Accelerating learning via knowledge transfer.
\newblock \emph{arXiv preprint arXiv:1511.05641}, 2015.

\bibitem[Chen et~al.(2016)Chen, Duan, Houthooft, Schulman, Sutskever, and
  Abbeel]{chen2016infogan}
Xi~Chen, Yan Duan, Rein Houthooft, John Schulman, Ilya Sutskever, and Pieter
  Abbeel.
\newblock Infogan: Interpretable representation learning by information
  maximizing generative adversarial nets.
\newblock \emph{Advances in neural information processing systems}, 29, 2016.

\bibitem[Cheng et~al.(2022)Cheng, Misra, Schwing, Kirillov, and
  Girdhar]{cheng2022masked}
Bowen Cheng, Ishan Misra, Alexander~G Schwing, Alexander Kirillov, and Rohit
  Girdhar.
\newblock Masked-attention mask transformer for universal image segmentation.
\newblock In \emph{Proceedings of the IEEE/CVF Conference on Computer Vision
  and Pattern Recognition}, pp.\  1290--1299, 2022.

\bibitem[Chiley et~al.(2022)Chiley, Thangarasa, Gupta, Samar, Hestness, and
  DeCoste]{chiley2022revbifpn}
Vitaliy Chiley, Vithursan Thangarasa, Abhay Gupta, Anshul Samar, Joel Hestness,
  and Dennis DeCoste.
\newblock Revbifpn: The fully reversible bidirectional feature pyramid network.
\newblock \emph{arXiv preprint arXiv:2206.14098}, 2022.

\bibitem[Dai et~al.(2017)Dai, Qi, Xiong, Li, Zhang, Hu, and
  Wei]{dai2017deformable}
Jifeng Dai, Haozhi Qi, Yuwen Xiong, Yi~Li, Guodong Zhang, Han Hu, and Yichen
  Wei.
\newblock Deformable convolutional networks.
\newblock In \emph{Proceedings of the IEEE international conference on computer
  vision}, pp.\  764--773, 2017.

\bibitem[Deng et~al.(2009)Deng, Dong, Socher, Li, Li, and
  Fei-Fei]{deng2009imagenet}
Jia Deng, Wei Dong, Richard Socher, Li-Jia Li, Kai Li, and Li~Fei-Fei.
\newblock Imagenet: A large-scale hierarchical image database.
\newblock In \emph{2009 IEEE conference on computer vision and pattern
  recognition}, pp.\  248--255. Ieee, 2009.

\bibitem[Desjardins et~al.(2012)Desjardins, Courville, and
  Bengio]{desjardins2012disentangling}
Guillaume Desjardins, Aaron Courville, and Yoshua Bengio.
\newblock Disentangling factors of variation via generative entangling.
\newblock \emph{arXiv preprint arXiv:1210.5474}, 2012.

\bibitem[Devlin et~al.(2018)Devlin, Chang, Lee, and Toutanova]{devlin2018bert}
Jacob Devlin, Ming-Wei Chang, Kenton Lee, and Kristina Toutanova.
\newblock Bert: Pre-training of deep bidirectional transformers for language
  understanding.
\newblock \emph{arXiv preprint arXiv:1810.04805}, 2018.

\bibitem[Ding et~al.(2021)Ding, Lian, Yang, Wang, Jin, Lu, and Luo]{ding2021hr}
Mingyu Ding, Xiaochen Lian, Linjie Yang, Peng Wang, Xiaojie Jin, Zhiwu Lu, and
  Ping Luo.
\newblock Hr-nas: Searching efficient high-resolution neural architectures with
  lightweight transformers.
\newblock In \emph{Proceedings of the IEEE/CVF Conference on Computer Vision
  and Pattern Recognition}, pp.\  2982--2992, 2021.

\bibitem[Ding et~al.(2022{\natexlab{a}})Ding, Zhang, Han, and
  Ding]{ding2022replknet}
Xiaohan Ding, Xiangyu Zhang, Jungong Han, and Guiguang Ding.
\newblock Scaling up your kernels to 31x31: Revisiting large kernel design in
  cnns.
\newblock In \emph{Proceedings of the IEEE/CVF Conference on Computer Vision
  and Pattern Recognition}, pp.\  11963--11975, 2022{\natexlab{a}}.

\bibitem[Ding et~al.(2022{\natexlab{b}})Ding, Zhang, Han, and
  Ding]{ding2022scaling}
Xiaohan Ding, Xiangyu Zhang, Jungong Han, and Guiguang Ding.
\newblock Scaling up your kernels to 31x31: Revisiting large kernel design in
  cnns.
\newblock In \emph{Proceedings of the IEEE/CVF Conference on Computer Vision
  and Pattern Recognition}, pp.\  11963--11975, 2022{\natexlab{b}}.

\bibitem[Dinh et~al.(2014)Dinh, Krueger, and Bengio]{dinh2014nice}
Laurent Dinh, David Krueger, and Yoshua Bengio.
\newblock Nice: Non-linear independent components estimation.
\newblock \emph{arXiv preprint arXiv:1410.8516}, 2014.

\bibitem[Dosovitskiy et~al.(2020)Dosovitskiy, Beyer, Kolesnikov, Weissenborn,
  Zhai, Unterthiner, Dehghani, Minderer, Heigold, Gelly,
  et~al.]{dosovitskiy2020image}
Alexey Dosovitskiy, Lucas Beyer, Alexander Kolesnikov, Dirk Weissenborn,
  Xiaohua Zhai, Thomas Unterthiner, Mostafa Dehghani, Matthias Minderer, Georg
  Heigold, Sylvain Gelly, et~al.
\newblock An image is worth 16x16 words: Transformers for image recognition at
  scale.
\newblock \emph{arXiv preprint arXiv:2010.11929}, 2020.

\bibitem[Fan et~al.(2021)Fan, Xiong, Mangalam, Li, Yan, Malik, and
  Feichtenhofer]{fan2021multiscale}
Haoqi Fan, Bo~Xiong, Karttikeya Mangalam, Yanghao Li, Zhicheng Yan, Jitendra
  Malik, and Christoph Feichtenhofer.
\newblock Multiscale vision transformers.
\newblock In \emph{Proceedings of the IEEE/CVF International Conference on
  Computer Vision}, pp.\  6824--6835, 2021.

\bibitem[Garau et~al.(2022)Garau, Bisagno, Sambugaro, and
  Conci]{garau2022interpretable}
Nicola Garau, Niccol{\`o} Bisagno, Zeno Sambugaro, and Nicola Conci.
\newblock Interpretable part-whole hierarchies and conceptual-semantic
  relationships in neural networks.
\newblock In \emph{Proceedings of the IEEE/CVF Conference on Computer Vision
  and Pattern Recognition}, pp.\  13689--13698, 2022.

\bibitem[Ghiasi et~al.(2019)Ghiasi, Lin, and Le]{ghiasi2019fpn}
Golnaz Ghiasi, Tsung-Yi Lin, and Quoc~V Le.
\newblock Nas-fpn: Learning scalable feature pyramid architecture for object
  detection.
\newblock In \emph{Proceedings of the IEEE/CVF conference on computer vision
  and pattern recognition}, pp.\  7036--7045, 2019.

\bibitem[Ghiasi et~al.(2021)Ghiasi, Zoph, Cubuk, Le, and Lin]{ghiasi2021multi}
Golnaz Ghiasi, Barret Zoph, Ekin~D Cubuk, Quoc~V Le, and Tsung-Yi Lin.
\newblock Multi-task self-training for learning general representations.
\newblock In \emph{Proceedings of the IEEE/CVF International Conference on
  Computer Vision}, pp.\  8856--8865, 2021.

\bibitem[Gomez et~al.(2017)Gomez, Ren, Urtasun, and
  Grosse]{gomez2017reversible}
Aidan~N Gomez, Mengye Ren, Raquel Urtasun, and Roger~B Grosse.
\newblock The reversible residual network: Backpropagation without storing
  activations.
\newblock \emph{Advances in neural information processing systems}, 30, 2017.

\bibitem[Han et~al.(2021)Han, Fan, Dai, Sun, Cheng, Liu, and
  Wang]{han2021connection}
Qi~Han, Zejia Fan, Qi~Dai, Lei Sun, Ming-Ming Cheng, Jiaying Liu, and Jingdong
  Wang.
\newblock On the connection between local attention and dynamic depth-wise
  convolution.
\newblock In \emph{International Conference on Learning Representations}, 2021.

\bibitem[He et~al.(2016)He, Zhang, Ren, and Sun]{he2016deep}
Kaiming He, Xiangyu Zhang, Shaoqing Ren, and Jian Sun.
\newblock Deep residual learning for image recognition.
\newblock In \emph{Proceedings of the IEEE conference on computer vision and
  pattern recognition}, pp.\  770--778, 2016.

\bibitem[He et~al.(2022)He, Chen, Xie, Li, Doll{\'a}r, and
  Girshick]{he2022masked}
Kaiming He, Xinlei Chen, Saining Xie, Yanghao Li, Piotr Doll{\'a}r, and Ross
  Girshick.
\newblock Masked autoencoders are scalable vision learners.
\newblock In \emph{Proceedings of the IEEE/CVF Conference on Computer Vision
  and Pattern Recognition}, pp.\  16000--16009, 2022.

\bibitem[Higgins et~al.(2017)Higgins, Matthey, Pal, Burgess, Glorot, Botvinick,
  Mohamed, and Lerchner]{higgins2017betavae}
Irina Higgins, Loic Matthey, Arka Pal, Christopher Burgess, Xavier Glorot,
  Matthew Botvinick, Shakir Mohamed, and Alexander Lerchner.
\newblock beta-{VAE}: Learning basic visual concepts with a constrained
  variational framework.
\newblock In \emph{International Conference on Learning Representations}, 2017.
\newblock URL \url{https://openreview.net/forum?id=Sy2fzU9gl}.

\bibitem[Hinton(2021)]{hinton2021represent}
Geoffrey Hinton.
\newblock How to represent part-whole hierarchies in a neural network.
\newblock \emph{arXiv preprint arXiv:2102.12627}, 2021.

\bibitem[Jacobsen et~al.(2018)Jacobsen, Smeulders, and
  Oyallon]{jacobsen2018revnet}
J{\"o}rn-Henrik Jacobsen, Arnold Smeulders, and Edouard Oyallon.
\newblock i-revnet: Deep invertible networks.
\newblock \emph{arXiv preprint arXiv:1802.07088}, 2018.

\bibitem[Jiang et~al.(2021)Jiang, Zhang, Hou, Cheng, and
  Wei]{jiang2021layercam}
Peng-Tao Jiang, Chang-Bin Zhang, Qibin Hou, Ming-Ming Cheng, and Yunchao Wei.
\newblock Layercam: Exploring hierarchical class activation maps for
  localization.
\newblock \emph{IEEE Transactions on Image Processing}, 30:\penalty0
  5875--5888, 2021.

\bibitem[Karras et~al.(2019)Karras, Laine, and Aila]{karras2019style}
Tero Karras, Samuli Laine, and Timo Aila.
\newblock A style-based generator architecture for generative adversarial
  networks.
\newblock In \emph{Proceedings of the IEEE/CVF conference on computer vision
  and pattern recognition}, pp.\  4401--4410, 2019.

\bibitem[Kolesnikov et~al.(2020)Kolesnikov, Beyer, Zhai, Puigcerver, Yung,
  Gelly, and Houlsby]{kolesnikov2020big}
Alexander Kolesnikov, Lucas Beyer, Xiaohua Zhai, Joan Puigcerver, Jessica Yung,
  Sylvain Gelly, and Neil Houlsby.
\newblock Big transfer (bit): General visual representation learning.
\newblock In \emph{European conference on computer vision}, pp.\  491--507.
  Springer, 2020.

\bibitem[Kornblith et~al.(2019)Kornblith, Norouzi, Lee, and Hinton]{cka}
Simon Kornblith, Mohammad Norouzi, Honglak Lee, and Geoffrey Hinton.
\newblock Similarity of neural network representations revisited.
\newblock In \emph{International Conference on Machine Learning}, pp.\
  3519--3529. PMLR, 2019.

\bibitem[Kulkarni et~al.(2015)Kulkarni, Whitney, Kohli, and
  Tenenbaum]{kulkarni2015deep}
Tejas~D Kulkarni, William~F Whitney, Pushmeet Kohli, and Josh Tenenbaum.
\newblock Deep convolutional inverse graphics network.
\newblock \emph{Advances in neural information processing systems}, 28, 2015.

\bibitem[Lee et~al.(2015)Lee, Xie, Gallagher, Zhang, and Tu]{lee2015deeply}
Chen-Yu Lee, Saining Xie, Patrick Gallagher, Zhengyou Zhang, and Zhuowen Tu.
\newblock Deeply-supervised nets.
\newblock In \emph{Artificial intelligence and statistics}, pp.\  562--570.
  PMLR, 2015.

\bibitem[Lillicrap et~al.(2020)Lillicrap, Santoro, Marris, Akerman, and
  Hinton]{lillicrap2020backpropagation}
Timothy~P Lillicrap, Adam Santoro, Luke Marris, Colin~J Akerman, and Geoffrey
  Hinton.
\newblock Backpropagation and the brain.
\newblock \emph{Nature Reviews Neuroscience}, 21\penalty0 (6):\penalty0
  335--346, 2020.

\bibitem[Lin et~al.(2014)Lin, Maire, Belongie, Hays, Perona, Ramanan,
  Doll{\'a}r, and Zitnick]{lin2014microsoft}
Tsung-Yi Lin, Michael Maire, Serge Belongie, James Hays, Pietro Perona, Deva
  Ramanan, Piotr Doll{\'a}r, and C~Lawrence Zitnick.
\newblock Microsoft coco: Common objects in context.
\newblock In \emph{European conference on computer vision}, pp.\  740--755.
  Springer, 2014.

\bibitem[Lin et~al.(2017)Lin, Doll{\'a}r, Girshick, He, Hariharan, and
  Belongie]{lin2017feature}
Tsung-Yi Lin, Piotr Doll{\'a}r, Ross Girshick, Kaiming He, Bharath Hariharan,
  and Serge Belongie.
\newblock Feature pyramid networks for object detection.
\newblock In \emph{Proceedings of the IEEE conference on computer vision and
  pattern recognition}, pp.\  2117--2125, 2017.

\bibitem[Liu et~al.(2019)Liu, Chen, Schroff, Adam, Hua, Yuille, and
  Fei-Fei]{liu2019auto}
Chenxi Liu, Liang-Chieh Chen, Florian Schroff, Hartwig Adam, Wei Hua, Alan~L
  Yuille, and Li~Fei-Fei.
\newblock Auto-deeplab: Hierarchical neural architecture search for semantic
  image segmentation.
\newblock In \emph{Proceedings of the IEEE/CVF conference on computer vision
  and pattern recognition}, pp.\  82--92, 2019.

\bibitem[Liu et~al.(2021)Liu, Lin, Cao, Hu, Wei, Zhang, Lin, and
  Guo]{liu2021swin}
Ze~Liu, Yutong Lin, Yue Cao, Han Hu, Yixuan Wei, Zheng Zhang, Stephen Lin, and
  Baining Guo.
\newblock Swin transformer: Hierarchical vision transformer using shifted
  windows.
\newblock In \emph{Proceedings of the IEEE/CVF International Conference on
  Computer Vision}, pp.\  10012--10022, 2021.

\bibitem[Liu et~al.(2022{\natexlab{a}})Liu, Hu, Lin, Yao, Xie, Wei, Ning, Cao,
  Zhang, Dong, et~al.]{liu2022swin}
Ze~Liu, Han Hu, Yutong Lin, Zhuliang Yao, Zhenda Xie, Yixuan Wei, Jia Ning, Yue
  Cao, Zheng Zhang, Li~Dong, et~al.
\newblock Swin transformer v2: Scaling up capacity and resolution.
\newblock In \emph{Proceedings of the IEEE/CVF Conference on Computer Vision
  and Pattern Recognition}, pp.\  12009--12019, 2022{\natexlab{a}}.

\bibitem[Liu et~al.(2022{\natexlab{b}})Liu, Mao, Wu, Feichtenhofer, Darrell,
  and Xie]{liu2022convnet}
Zhuang Liu, Hanzi Mao, Chao-Yuan Wu, Christoph Feichtenhofer, Trevor Darrell,
  and Saining Xie.
\newblock A convnet for the 2020s.
\newblock In \emph{Proceedings of the IEEE/CVF Conference on Computer Vision
  and Pattern Recognition}, pp.\  11976--11986, 2022{\natexlab{b}}.

\bibitem[Locatello et~al.(2019)Locatello, Bauer, Lucic, Raetsch, Gelly,
  Sch{\"o}lkopf, and Bachem]{locatello2019challenging}
Francesco Locatello, Stefan Bauer, Mario Lucic, Gunnar Raetsch, Sylvain Gelly,
  Bernhard Sch{\"o}lkopf, and Olivier Bachem.
\newblock Challenging common assumptions in the unsupervised learning of
  disentangled representations.
\newblock In \emph{international conference on machine learning}, pp.\
  4114--4124. PMLR, 2019.

\bibitem[Ma et~al.(2020)Ma, Zhang, Huang, and Sun]{ma2020weightnet}
Ningning Ma, Xiangyu Zhang, Jiawei Huang, and Jian Sun.
\newblock Weightnet: Revisiting the design space of weight networks.
\newblock In \emph{European Conference on Computer Vision}, pp.\  776--792.
  Springer, 2020.

\bibitem[Mahajan et~al.(2018)Mahajan, Girshick, Ramanathan, He, Paluri, Li,
  Bharambe, and Van Der~Maaten]{mahajan2018exploring}
Dhruv Mahajan, Ross Girshick, Vignesh Ramanathan, Kaiming He, Manohar Paluri,
  Yixuan Li, Ashwin Bharambe, and Laurens Van Der~Maaten.
\newblock Exploring the limits of weakly supervised pretraining.
\newblock In \emph{Proceedings of the European conference on computer vision
  (ECCV)}, pp.\  181--196, 2018.

\bibitem[Mangalam et~al.(2022)Mangalam, Fan, Li, Wu, Xiong, Feichtenhofer, and
  Malik]{mangalam2022reversible}
Karttikeya Mangalam, Haoqi Fan, Yanghao Li, Chao-Yuan Wu, Bo~Xiong, Christoph
  Feichtenhofer, and Jitendra Malik.
\newblock Reversible vision transformers.
\newblock In \emph{Proceedings of the IEEE/CVF Conference on Computer Vision
  and Pattern Recognition}, pp.\  10830--10840, 2022.

\bibitem[Oord et~al.(2018)Oord, Li, and Vinyals]{oord2018representation}
Aaron van~den Oord, Yazhe Li, and Oriol Vinyals.
\newblock Representation learning with contrastive predictive coding.
\newblock \emph{arXiv preprint arXiv:1807.03748}, 2018.

\bibitem[Ott et~al.(2018)Ott, Edunov, Grangier, and
  Auli]{ott-etal-2018-scaling}
Myle Ott, Sergey Edunov, David Grangier, and Michael Auli.
\newblock Scaling neural machine translation.
\newblock In \emph{Proceedings of the Third Conference on Machine Translation:
  Research Papers}, pp.\  1--9, Brussels, Belgium, October 2018. Association
  for Computational Linguistics.
\newblock \doi{10.18653/v1/W18-6301}.

\bibitem[Radford et~al.(2021)Radford, Kim, Hallacy, Ramesh, Goh, Agarwal,
  Sastry, Askell, Mishkin, Clark, et~al.]{radford2021learningclip}
Alec Radford, Jong~Wook Kim, Chris Hallacy, Aditya Ramesh, Gabriel Goh,
  Sandhini Agarwal, Girish Sastry, Amanda Askell, Pamela Mishkin, Jack Clark,
  et~al.
\newblock Learning transferable visual models from natural language
  supervision.
\newblock In \emph{International Conference on Machine Learning}, pp.\
  8748--8763. PMLR, 2021.

\bibitem[Ridnik et~al.(2021)Ridnik, Ben-Baruch, Noy, and
  Zelnik-Manor]{ridnik2021imagenet}
Tal Ridnik, Emanuel Ben-Baruch, Asaf Noy, and Lihi Zelnik-Manor.
\newblock Imagenet-21k pretraining for the masses.
\newblock \emph{arXiv preprint arXiv:2104.10972}, 2021.

\bibitem[Ruder(2017)]{ruder2017overview}
Sebastian Ruder.
\newblock An overview of multi-task learning in deep neural networks.
\newblock \emph{arXiv preprint arXiv:1706.05098}, 2017.

\bibitem[Sener \& Koltun(2018)Sener and Koltun]{sener2018multi}
Ozan Sener and Vladlen Koltun.
\newblock Multi-task learning as multi-objective optimization.
\newblock \emph{Advances in neural information processing systems}, 31, 2018.

\bibitem[Sennrich et~al.(2016)Sennrich, Haddow, and Birch]{sennrich2016neural}
Rico Sennrich, Barry Haddow, and Alexandra Birch.
\newblock Neural machine translation of rare words with subword units.
\newblock In \emph{54th Annual Meeting of the Association for Computational
  Linguistics}, pp.\  1715--1725. Association for Computational Linguistics
  (ACL), 2016.

\bibitem[Shao et~al.(2019)Shao, Li, Zhang, Peng, Yu, Zhang, Li, and
  Sun]{shao2019objects365}
Shuai Shao, Zeming Li, Tianyuan Zhang, Chao Peng, Gang Yu, Xiangyu Zhang, Jing
  Li, and Jian Sun.
\newblock Objects365: A large-scale, high-quality dataset for object detection.
\newblock In \emph{Proceedings of the IEEE/CVF international conference on
  computer vision}, pp.\  8430--8439, 2019.

\bibitem[Szegedy et~al.(2015)Szegedy, Liu, Jia, Sermanet, Reed, Anguelov,
  Erhan, Vanhoucke, and Rabinovich]{szegedy2015going}
Christian Szegedy, Wei Liu, Yangqing Jia, Pierre Sermanet, Scott Reed, Dragomir
  Anguelov, Dumitru Erhan, Vincent Vanhoucke, and Andrew Rabinovich.
\newblock Going deeper with convolutions.
\newblock In \emph{Proceedings of the IEEE conference on computer vision and
  pattern recognition}, pp.\  1--9, 2015.

\bibitem[Tan \& Le(2019)Tan and Le]{tan2019efficientnet}
Mingxing Tan and Quoc Le.
\newblock Efficientnet: Rethinking model scaling for convolutional neural
  networks.
\newblock In \emph{International conference on machine learning}, pp.\
  6105--6114. PMLR, 2019.

\bibitem[Tan et~al.(2020)Tan, Pang, and Le]{tan2020efficientdet}
Mingxing Tan, Ruoming Pang, and Quoc~V Le.
\newblock Efficientdet: Scalable and efficient object detection.
\newblock In \emph{Proceedings of the IEEE/CVF conference on computer vision
  and pattern recognition}, pp.\  10781--10790, 2020.

\bibitem[Thomee et~al.(2016)Thomee, Shamma, Friedland, Elizalde, Ni, Poland,
  Borth, and Li]{thomee2016yfcc100m}
Bart Thomee, David~A Shamma, Gerald Friedland, Benjamin Elizalde, Karl Ni,
  Douglas Poland, Damian Borth, and Li-Jia Li.
\newblock Yfcc100m: The new data in multimedia research.
\newblock \emph{Communications of the ACM}, 59\penalty0 (2):\penalty0 64--73,
  2016.

\bibitem[Tishby \& Zaslavsky(2015)Tishby and Zaslavsky]{tishby2015deep}
Naftali Tishby and Noga Zaslavsky.
\newblock Deep learning and the information bottleneck principle.
\newblock In \emph{2015 ieee information theory workshop (itw)}, pp.\  1--5.
  IEEE, 2015.

\bibitem[Tishby et~al.(2000)Tishby, Pereira, and Bialek]{tishby2000information}
Naftali Tishby, Fernando~C Pereira, and William Bialek.
\newblock The information bottleneck method.
\newblock \emph{arXiv preprint physics/0004057}, 2000.

\bibitem[Touvron et~al.(2020)Touvron, Cord, Douze, Massa, Sablayrolles, and
  J{\'e}gou]{touvron2020training}
Hugo Touvron, Matthieu Cord, Matthijs Douze, Francisco Massa, Alexandre
  Sablayrolles, and Herv{\'e} J{\'e}gou.
\newblock Training data-efficient image transformers \& distillation through
  attention.
\newblock \emph{arXiv preprint arXiv:2012.12877}, 2020.

\bibitem[Vaswani et~al.(2017)Vaswani, Shazeer, Parmar, Uszkoreit, Jones, Gomez,
  Kaiser, and Polosukhin]{vaswani2017attention}
Ashish Vaswani, Noam Shazeer, Niki Parmar, Jakob Uszkoreit, Llion Jones,
  Aidan~N Gomez, {\L}ukasz Kaiser, and Illia Polosukhin.
\newblock Attention is all you need.
\newblock \emph{Advances in neural information processing systems}, 30, 2017.

\bibitem[Wang et~al.(2020)Wang, Sun, Cheng, Jiang, Deng, Zhao, Liu, Mu, Tan,
  Wang, et~al.]{wang2020deep}
Jingdong Wang, Ke~Sun, Tianheng Cheng, Borui Jiang, Chaorui Deng, Yang Zhao,
  Dong Liu, Yadong Mu, Mingkui Tan, Xinggang Wang, et~al.
\newblock Deep high-resolution representation learning for visual recognition.
\newblock \emph{IEEE transactions on pattern analysis and machine
  intelligence}, 43\penalty0 (10):\penalty0 3349--3364, 2020.

\bibitem[Wang et~al.(2022)Wang, Bao, Dong, Bjorck, Peng, Liu, Aggarwal,
  Mohammed, Singhal, Som, et~al.]{wang2022beit3}
Wenhui Wang, Hangbo Bao, Li~Dong, Johan Bjorck, Zhiliang Peng, Qiang Liu, Kriti
  Aggarwal, Owais~Khan Mohammed, Saksham Singhal, Subhojit Som, et~al.
\newblock Image as a foreign language: Beit pretraining for all vision and
  vision-language tasks.
\newblock \emph{arXiv preprint arXiv:2208.10442}, 2022.

\bibitem[Wang et~al.(2021)Wang, Ni, Song, Yang, and Huang]{wang2021revisiting}
Yulin Wang, Zanlin Ni, Shiji Song, Le~Yang, and Gao Huang.
\newblock Revisiting locally supervised learning: an alternative to end-to-end
  training.
\newblock \emph{arXiv preprint arXiv:2101.10832}, 2021.

\bibitem[Wu et~al.(2021)Wu, Li, Zhang, Dai, Zhang, Yu, Wang, Lin, and
  Vajda]{wu2021fbnetv5}
Bichen Wu, Chaojian Li, Hang Zhang, Xiaoliang Dai, Peizhao Zhang, Matthew Yu,
  Jialiang Wang, Yingyan Lin, and Peter Vajda.
\newblock Fbnetv5: Neural architecture search for multiple tasks in one run.
\newblock \emph{arXiv preprint arXiv:2111.10007}, 2021.

\bibitem[Xiao et~al.(2018)Xiao, Liu, Zhou, Jiang, and Sun]{xiao2018unified}
Tete Xiao, Yingcheng Liu, Bolei Zhou, Yuning Jiang, and Jian Sun.
\newblock Unified perceptual parsing for scene understanding.
\newblock In \emph{Proceedings of the European conference on computer vision
  (ECCV)}, pp.\  418--434, 2018.

\bibitem[Xie et~al.(2022)Xie, Zhang, Cao, Lin, Bao, Yao, Dai, and
  Hu]{xie2022simmim}
Zhenda Xie, Zheng Zhang, Yue Cao, Yutong Lin, Jianmin Bao, Zhuliang Yao,
  Qi~Dai, and Han Hu.
\newblock Simmim: A simple framework for masked image modeling.
\newblock In \emph{Proceedings of the IEEE/CVF Conference on Computer Vision
  and Pattern Recognition}, pp.\  9653--9663, 2022.

\bibitem[Yalniz et~al.(2019)Yalniz, J{\'e}gou, Chen, Paluri, and
  Mahajan]{yalniz2019billion}
I~Zeki Yalniz, Herv{\'e} J{\'e}gou, Kan Chen, Manohar Paluri, and Dhruv
  Mahajan.
\newblock Billion-scale semi-supervised learning for image classification.
\newblock \emph{arXiv preprint arXiv:1905.00546}, 2019.

\bibitem[Yuan et~al.(2021{\natexlab{a}})Yuan, Chen, Wang, Yu, Shi, Jiang, Tay,
  Feng, and Yan]{yuan2021tokens}
Li~Yuan, Yunpeng Chen, Tao Wang, Weihao Yu, Yujun Shi, Zi-Hang Jiang,
  Francis~EH Tay, Jiashi Feng, and Shuicheng Yan.
\newblock Tokens-to-token vit: Training vision transformers from scratch on
  imagenet.
\newblock In \emph{Proceedings of the IEEE/CVF International Conference on
  Computer Vision}, pp.\  558--567, 2021{\natexlab{a}}.

\bibitem[Yuan et~al.(2021{\natexlab{b}})Yuan, Chen, Chen, Codella, Dai, Gao,
  Hu, Huang, Li, Li, et~al.]{yuan2021florence}
Lu~Yuan, Dongdong Chen, Yi-Ling Chen, Noel Codella, Xiyang Dai, Jianfeng Gao,
  Houdong Hu, Xuedong Huang, Boxin Li, Chunyuan Li, et~al.
\newblock Florence: A new foundation model for computer vision.
\newblock \emph{arXiv preprint arXiv:2111.11432}, 2021{\natexlab{b}}.

\bibitem[Zamir et~al.(2018)Zamir, Sax, Shen, Guibas, Malik, and
  Savarese]{zamir2018taskonomy}
Amir~R Zamir, Alexander Sax, William Shen, Leonidas~J Guibas, Jitendra Malik,
  and Silvio Savarese.
\newblock Taskonomy: Disentangling task transfer learning.
\newblock In \emph{Proceedings of the IEEE conference on computer vision and
  pattern recognition}, pp.\  3712--3722, 2018.

\bibitem[Zhang et~al.(2022{\natexlab{a}})Zhang, Li, Liu, Zhang, Su, Zhu, Ni,
  and Shum]{zhang2022dino}
Hao Zhang, Feng Li, Shilong Liu, Lei Zhang, Hang Su, Jun Zhu, Lionel~M Ni, and
  Heung-Yeung Shum.
\newblock Dino: Detr with improved denoising anchor boxes for end-to-end object
  detection.
\newblock \emph{arXiv preprint arXiv:2203.03605}, 2022{\natexlab{a}}.

\bibitem[Zhang et~al.(2022{\natexlab{b}})Zhang, Sun, Zhou, He, Yin, Wang,
  Sheng, Qiao, Shao, and Liu]{zhang2022bamboo}
Yuanhan Zhang, Qinghong Sun, Yichun Zhou, Zexin He, Zhenfei Yin, Kun Wang,
  Lu~Sheng, Yu~Qiao, Jing Shao, and Ziwei Liu.
\newblock Bamboo: Building mega-scale vision dataset continually with
  human-machine synergy, 2022{\natexlab{b}}.

\bibitem[Zhou et~al.(2017{\natexlab{a}})Zhou, Lapedriza, Khosla, Oliva, and
  Torralba]{zhou2017places}
Bolei Zhou, Agata Lapedriza, Aditya Khosla, Aude Oliva, and Antonio Torralba.
\newblock Places: A 10 million image database for scene recognition.
\newblock \emph{IEEE Transactions on Pattern Analysis and Machine
  Intelligence}, 2017{\natexlab{a}}.

\bibitem[Zhou et~al.(2017{\natexlab{b}})Zhou, Zhao, Puig, Fidler, Barriuso, and
  Torralba]{zhou2017scene}
Bolei Zhou, Hang Zhao, Xavier Puig, Sanja Fidler, Adela Barriuso, and Antonio
  Torralba.
\newblock Scene parsing through ade20k dataset.
\newblock In \emph{Proceedings of the IEEE conference on computer vision and
  pattern recognition}, pp.\  633--641, 2017{\natexlab{b}}.

\end{thebibliography}
\bibliographystyle{iclr2023_conference}

\clearpage
\appendix
\section{Micro Design Details}
\label{appendix:microdesign}
\begin{figure}[h]
\begin{center}
\includegraphics[width=1.0\columnwidth]{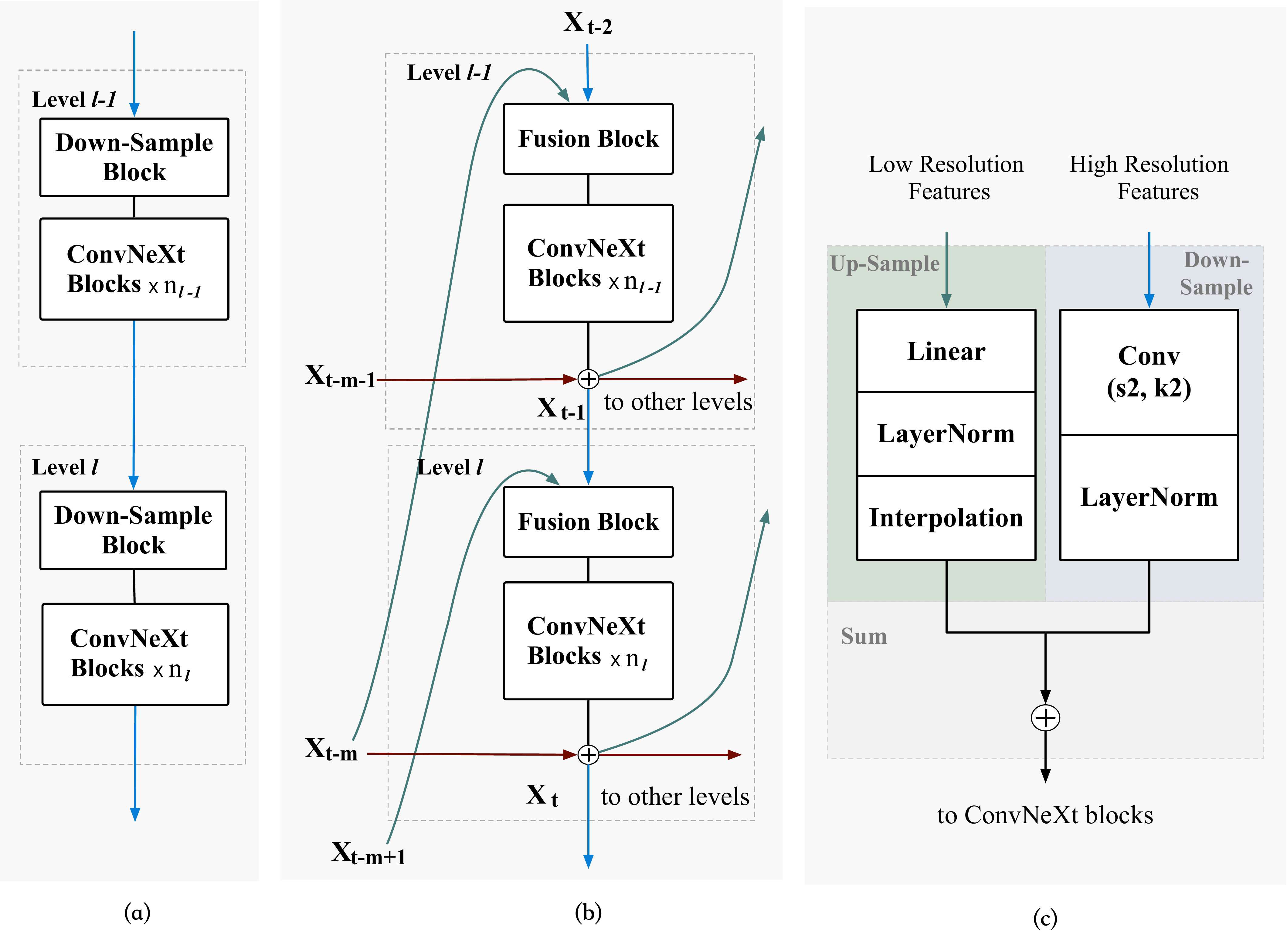}
\end{center}

\caption{(a) Levels in ConvNeXt. Level $l$ contains a patch merging down-sample block and $n_{l}$ residual blocks. (b) Levels in \net. Level $l$ is composed of a fusion module, $n_{l}$ residual blocks and a reversible operation. Note that Level $l$ takes features maps $x_{t-1}$, $x_{t-m+1}$ and $x_{t-m}$ as input. Feature maps $x_{t-1}$ and $x_{t-m+1}$ are fed into the fusion module and feature maps $x_{t-m}$ are fed into the reversible operation. (c) Design of the fusion module. }

\label{fig:micro}
\end{figure}
In this section, we provide the architecture design details for \net. 
As depicted in Fig. \ref{fig:arch} and Section \ref{3:arch}, our \net~contains multiple columns with reversible connections. Fig. \ref{fig:micro} (a) shows the architecture of ConvNeXt. Note that we replace the $7\times7$ depth-wise convolution in ConvNeXt with $3\times3$, as described in Sec.~\ref{sec:micro}. In Fig.~\ref{fig:micro}~(b), we show in detail how to extend to our \net~on the basis of ConvNeXt. First, we replace the down-sample block with a fusion block to fuse low-level representations in current column and high-level ones from the previous column, and Fig.~\ref{fig:micro}~(c) shows the details of fusion block which contains up-sample and down-sample operations to handle different resolutions. Second, for each level, same-level representations from the previous column are added to current level's output and are ready to propagate as a whole. Thanks to the two modifications, feature maps from different hierarchies aggregate together to form the intermediate representation. In Fig.~\ref{fig:micro}~(c), we use a \texttt{Linear-LayerNorm} followed by a nearest interpolation to up-sample low resolution features. A $2\times 2$ kernel \texttt{Conv2d} with stride 2 down-samples the high resolution features, followed by a \texttt{LayerNorm} to balance the contributions of the two inputs.

\section{Generalization to Transformers}
\label{appendix:transformer}
\subsection{Vision Transformer Models}
\label{appendix:transformer-vit}
{\net~contains multiple light-weight sub-networks
with reversible connections. 
In this paper, we adopt the ConvNext micro design by default
except for multi-columns fusion and smaller convolution kernel
as described in Sec. \ref{sec:micro}. 
However, the micro design of our \net~is not limited to convolutional networks, 
but is also compatible with isotropic designing, 
such as the vanilla vision transformer (ViT)~\citep{dosovitskiy2020image}.
In this section, we show the micro design of \net~can generalized to vanilla ViT,
namely \net-ViT,
with promising experimental results.}

net-ViT maintains the feature resolution 
in the reversible columns.
Thus the patch merging blocks and up-sample blocks in the 
fusion modules are replaced with a simple linear projection with
a post LayerNorm. We use the vanilla ViT building block
instead of the ConvNext building block variant. 
The post LayerNorms and normalized dot-product attention are used in ViT blocks to stabilize training convergence, similar to~\cite{liu2022swin}. With the properties of isotropy, we evenly arrange the building blocks in each column. The configuration details of \net-ViT are:

\begin{itemize}
 \small
  \item \net-ViT-S: $C = (224, 224, 224, 224)$, $B = (2, 2, 2, 2)$, $\textit{HEAD} = 4$, $\textit{COL} = 4$
  \item \net-ViT-B: $C = (384, 384, 384, 384)$, $B = (3, 3, 3, 3)$, $\textit{HEAD} = 6$, $\textit{COL} = 4$
\end{itemize}

\begin{table}[h]
\small
\setlength{\tabcolsep}{5pt}
\centering
\caption{ImageNet-1K classification results. 
We compare our \net-ViT with
state-of-the-art isotropic \bv Vision Transformers
and \bc CNNs that have comparable FLOPs and parameters.}
\label{tab:isotropic_imagenet}
\vspace{0.1cm}
\begin{tabu}{lcccc}
\toprule

Model & Image Size& Params & FLOPs & Top-1 Acc. \\ 
\midrule
\bv DeiT-S~\citep{touvron2020training} & $224^2$ & 22M & 4.6G & 79.8 \\
\bc ConvNext-S (\textit{iso.})~\citep{liu2022convnet} & $224^2$ & 22M & 4.3G & 79.7 \\
\bv \textbf{\net-ViT-S} & $224^2$ & 16M & 4.6G & \textbf{80.6} \\
\hline 
\bv ViT-B~\citep{dosovitskiy2020image} & $384^2$ & 86M & 55.4G & 77.9 \\
\bv DeiT-B~\citep{touvron2020training} & $224^2$ & 86M & 17.6G & 81.7 \\
\bv Rev-ViT-B~\citep{mangalam2022reversible} & $224^2$ & 87M & 17.6G & 81.8 \\
\bv Rev-MViT-B~\citep{mangalam2022reversible} & $224^2$ & 39M & 8.7G & 82.5 \\
\bc ConvNext-B (\textit{iso.})~\citep{liu2022convnet} & $224^2$ & 87M & 16.9G & 82.0 \\
\bv \textbf{\net-ViT-B} & $224^2$ & 67M & 18.8G & \textbf{82.7} \\

\bottomrule
\end{tabu}%
\end{table}

We use the same training setting with the anisotropic \net~as described in Sec.~\ref{chap:img},
except that the intermediate supervision is 
discarded for simplicity and the stochastic depth rate is set as 0.2 for 
\net-B.
We scale down the value of last linear projection layers in each FFN accroding to the network depth in initialization, same as BEiT~\citep{bao2021beit}. 
In Tab.~\ref{tab:isotropic_imagenet}, we compare the \net-ViT
with vanilla ViT and other concurrent isotropic designs. 
Our \net-ViT 
surpasses vanilla vision transformer (77.9\% for ViT and 81.7\% for DeiT) 
and convolutional network ConvNeXt (82.0\%) 
that have comparable model parameters and 
computational overhead
on ImageNet-1k classification \wrt the top-1 accuracy.

\subsection{Language Models}
\label{appendix:nlp}

Considering the great success of applying transformer to computer vision, i.e., ViT~\citep{dosovitskiy2020image},
we also made some exploration to generalize RevCol to natural language processing (NLP).
Based on the design in Appendix~\ref{appendix:transformer-vit}, we can easily apply the isotropic RevCol to language models with minor modification.
To be specific, we replace the stem module in our RevCol with word embedding and positional encoding in transformer.
Then, the RevCol can be plugged into the original transformer as an encoder.
The output of the last column in RevCol will be used as the memory keys and values for the attention layers in decoder, just exactly the same as the original transformer.

We select the translation task to evaluate the potential of the RevCol in NLP.
We run experiments on the WMT'16 English-German (En-De) dataset with 4.5M sentences and larger WMT'14 English-French dataset with 36M sentences.
Each sentence is encoded by joint source and target byte pair encoding following~\cite{sennrich2016neural}.
The details of model architecture and the BLEU score are shown in Tab.~\ref{tab:translation_wmt16}.

\begin{table}[h]
\small
\setlength{\tabcolsep}{5pt}
\centering
\caption{BLEU score on newstest2014 for WMT English-German (En-De) and English-French (En-Fr) translation task. $^\dagger$ indicates we re-run the experiments with \texttt{fairseq}.}
\label{tab:translation_wmt16}
\vspace{0.1cm}
\adjustbox{width=\linewidth}{
\begin{tabu}{lcccccccccccc}
\toprule
\multirow{2}{*}{Model} & \multicolumn{4}{c}{Encoder}& \ & \multicolumn{4}{c}{Decoder} &\multirow{2}{*}{Params}  & \multirow{2}{*}{Task}&\multirow{2}{*}{BLEU} \\
\cmidrule(lr){2-5}\cmidrule(lr){7-10}
& arch   &  d$_{model}$  &  d$_{ff}$ & head && arch  &  d$_{model}$  &  d$_{ff}$ & head\\
\hline
Transformer$_{big}^\dagger$ & \multirow{2}{*}{\textit{N} = 6} &   \multirow{2}{*}{1024}&   \multirow{2}{*}{4096}    &\multirow{2}{*}{16} &&\multirow{2}{*}{\textit{N} = 6}&   \multirow{2}{*}{1024}&   \multirow{2}{*}{4096}    &\multirow{2}{*}{16}&   209M & En-De &  28.43\\
\citep{vaswani2017attention}&&&&&&&&&&221M&En-Fr&43.07\\
\hline
\noalign{\vskip 0.1cm}
\multirow{2}{*}{\textbf{RevCol-Transformer}} &\textit{B} = (1,1,1,1) & \multirow{2}{*}{768}&   \multirow{2}{*}{3072}    &\multirow{2}{*}{12} &&\multirow{2}{*}{\textit{N} = 6}&   \multirow{2}{*}{768}&   \multirow{2}{*}{3072}    &\multirow{2}{*}{12} & 200M&  En-De    &   \textbf{28.67}\\
&\textit{COL} = 4&&&&&&&&&209M&En-Fr&\textbf{43.40}\\
\bottomrule
\end{tabu}}
\end{table}

All the dataset preparation and the training configurations follows~\cite{ott-etal-2018-scaling} and the open source project \texttt{fairseq}.
The models were trained for 300K steps with batch-size of 28,672 tokens on En-De and 200K steps with batch-size of 86,016 on En-Fr.
We discard the intermediate supervision for simplicity.
As shown in Tab.~\ref{tab:translation_wmt16}, our RevCol outperforms vanilla transformer with comparable parameters on En-De (28.67 \vs 28.43) and En-Fr (43.40 \vs 43.07), which demonstrates the RevCol's applicability to NLP.

\subsection{Robustness of the number of columns}

In the ablation analysis of the paper, we show that when fix the total FLOPs and add more columns of \net, the performance first increases and then get saturated. When the number of columns is extreme large, such as 20, the performance drop because of the representation ability of single column is limited. When the number of columns is usual, such as 4\~12, the performances are similar, which verifies the setting robustness of the number of columns.

\begin{figure}[ht]
    \centering
    \includegraphics[width=0.6\textwidth]{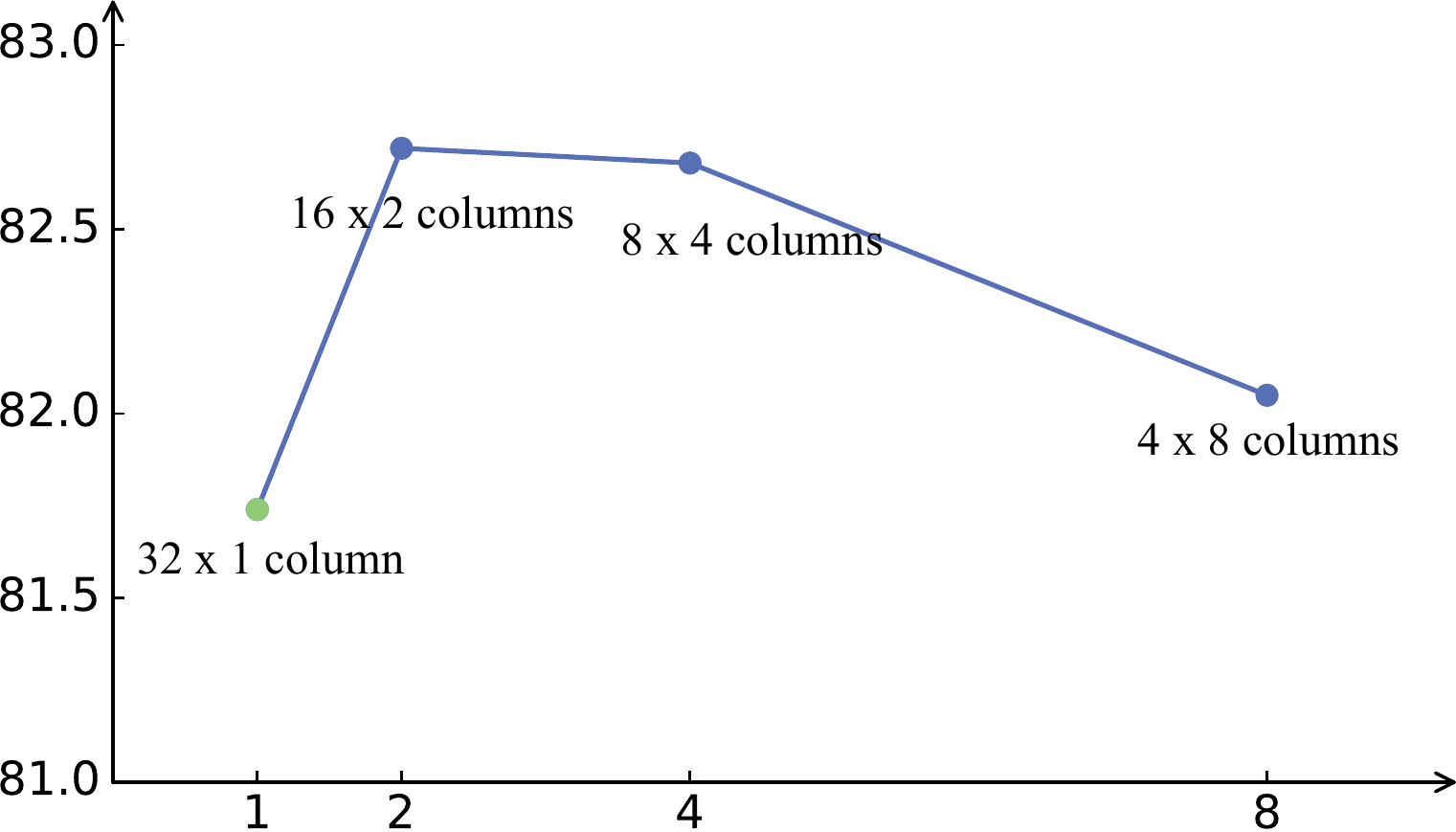}
    \caption{ImageNet top-1 accuracy of different variants of \net-ViT-B. Each variant has the same total number of residual blocks and channel dimension.}
    \label{fig:fix_blocks}
\end{figure}

To further analyze the robustness of the number of columns, in this section, we build some \net-ViT-B variants (see Appendix~\ref{appendix:transformer} for more details). Each variant has the same number of residual blocks with the same channel dimension, but different number of columns. In other worlds, these variants have the same channel dimension and different depth of each columns and different number of columns. 
We use 32 residual blocks totally and maintain the FLOPs about 18G. Fig.~\ref{fig:fix_blocks} show the performance on ImageNet-1K of different variants. The number of columns are 1, 2, 4, and 8, accordingly the depth of each column are 32, 16, 8, and 4. The performance of single column variant is lower (similar to DeiT-B~\citep{touvron2020training}) because of the single column ViT can not maintain the information as multi reversible columns.
The performance is decreasing when the number of columns became larger, because of the depth of each columns is not enough. This phenomenon indicates us that given a target FLOPs, the setting of the number of columns is robust unless the depth of each columns or channel dimension is too small.

\section{Semi-labeled Privately Collected Dataset for Large Models}
\label{appendix:external-dataset}

\subsection{Data Collection and Pseudo Label System}
  The dataset consists of around 168 million(M) images, 50M of which labeled and the remaining 118M unlabeled. The majority of labeled images come from public datasets, \eg ImageNet, Places365~\citep{zhou2017places}, and Bamboo~\citep{zhang2022bamboo}. The others are web-crawled images annotated by in-door employees. Unlabeled images come from weakly-annotated image-text datasets like YFCC-100M~\citep{thomee2016yfcc100m}. We do not use text annotations.

  In order to utilize images of different label domains and the massive unlabeled images, we employ a multi-target label system similar to \cite{ding2022replknet} and \cite{ghiasi2021multi}. We adopt a semi-supervised learning strategy with ViTs, thus generating \textit{pseudo} labels with continuously increased quality. We only store soft predictions with confidence higher than 1\% to save storage. The final version of \textit{pseudo} label we use are generated by a multi-head ViT-Huge teacher, which has an 89.0\% ImageNet-1k accuracy.

\subsection{Image Deduplication}

    Since the dataset contains large amount of unverified web-crawled images, there are probably validation or test images sneaking into our training dataset. Works like 
\cite{mahajan2018exploring} and \cite{yalniz2019billion} all regard image deduplication an important procedure for fair experiments.

\begin{figure}[h]
\begin{center}

\includegraphics[width=1.0\columnwidth]{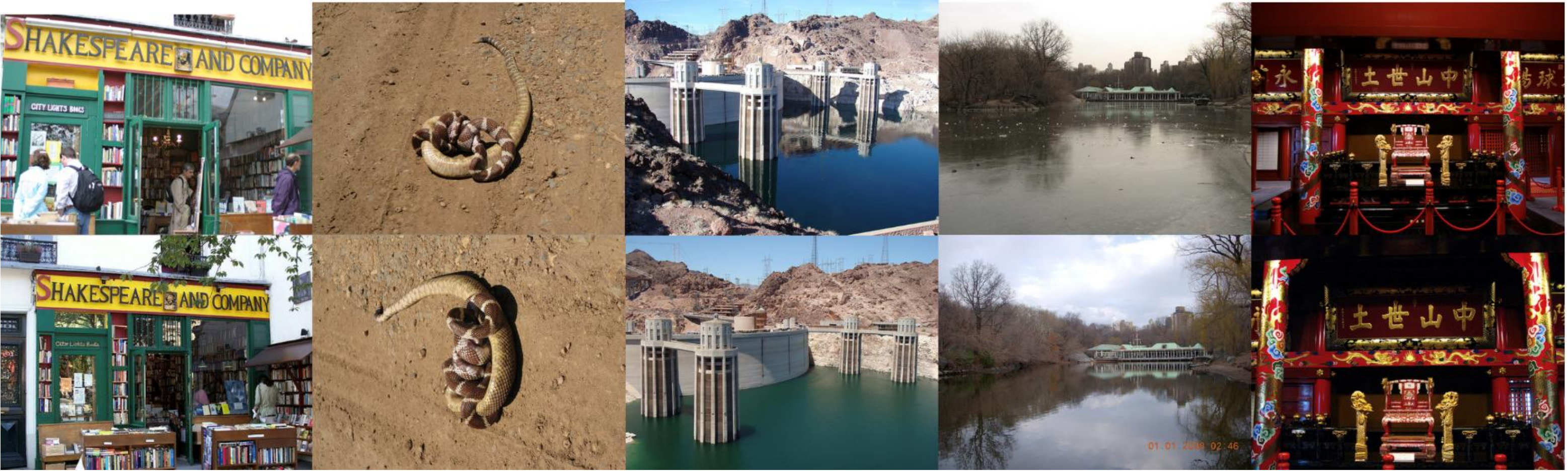}
\end{center}
\caption{Top: Near duplicates found in unlabeled images. Bottom: ImageNet-1k validation images.}
\label{fig:near-duplicates}
\end{figure}

  We first iterate over the entire dataset to filter out suspicious duplicates together with the corresponding test images based on their \textit{pseudo} label distance. This brings more than 10,000 images with high suspicion. We look at these image pairs and finally find about 1,200 exact-duplicates and near-duplicates. Fig.~\ref{fig:near-duplicates} shows some examples of the near-duplicates, which are difficult to detect. Never the less, training a model without removing these duplicates gives less than 0.1\% accuracy gain on ImageNet-1k in our experiments. We attribute this to the absence of true labels from these duplicates.

\section{More Training Details} \label{appendix:details}

This section gives more training details on ImageNet classification, COCO detection, and ADE20K segmentation.

\subsection{Intermediate Supervision Settings}
We add intermediate supervision in ImageNet-1k training, ImageNet-22k and extra data pre-training. We used a 3-block decoder with gradually up-sampled feature maps in ImageNet-1k training. The block setting remains the same as Sec. \ref{3:arch} We use a single layer decoder in ImageNet-22k and extra data pre-training. For all the variants of \net, we set the number of compound loss $n$ to 3 empirically (eg. for a 8 column \net, the intermediate supervision is added to column 2, 4, and 6, and the original classification CE loss is also added to column 8). $\alpha_i$ is set to {3, 2, 1, 0} and $\beta_{i}$ is set to {0.18, 0.35, 0.53, 1}.
\subsection{Hyperparameters Used for Training and Pre-training}

This section introduces the training details for main experiments, the supervised training on ImageNet and extra data. We show this setting in Tab. \ref{tbl:pretrain:hyperparams}. 
All experiments in ablation studies are superivised trained on ImageNet-1K except additional descriptions and also follow settings described in this section.

\begin{table}[H]
\centering
\small
\caption{
Hyperparameters for training and pre-training \net.
}
\begin{tabular}{l|cc|c}
\toprule

\multirow{2}[2]{*}{\bf Hyperparameters} & \bf ImageNet-1K & \bf ImageNet-22K & \bf 168M Extra Data \\
\cmidrule(lr){2-2}  \cmidrule(lr){3-3}\cmidrule(lr){4-4} 
& \bf T/S/B & \bf B/L/XL & \bf XL/H\\
\midrule
Input resolution & \multicolumn{2}{c|}{$224^2$} & $224^2$ \\
Training epochs & 300 & 90 & 10 \\
Warmup epochs & 20 & 5 & 0.15 \\
Batch size & \multicolumn{2}{c|}{4096} & 5120 \\
Peak learning rate & 4e-3 & 5e-4 & 6.25e-4 \\
Learning rate schedule & \multicolumn{2}{c|}{cosine} & cosine \\
Layer-wise learning rate decay &\multicolumn{2}{c|}{\xmark} &\xmark \\
AdamW momentum & \multicolumn{2}{c|}{(0.9, 0.999)} & (0.9, 0.999) \\
Weight decay & 0.05 & 0.1 & 0.05 \\
Gradient clipping & \multicolumn{2}{c|}{\xmark} & 1.0 (element-wise) \\
Drop path & 0.1/0.3/0.4 & 0.3 & 0.2 \\
EMA & 0.9999 & \xmark &\xmark\\
\midrule
Label smoothing $\varepsilon$ & \multicolumn{2}{c|}{0.1} & 0.1 \\
Data augment & \multicolumn{2}{c|}{RandAug (9, 0.5)} & RandAug (9, 0.5) \\
Mixup & \multicolumn{2}{c|}{0.8} & \xmark \\
CutMix & \multicolumn{2}{c|}{1.0} & \xmark \\
Random erase & \multicolumn{2}{c|}{0.25} & \xmark \\
\bottomrule
\end{tabular}

\label{tbl:pretrain:hyperparams}
\end{table}

\subsection{Hyperparameters Used for Fine-tuning}
This section gives the hyperparameters used for fine-tuning on ImageNet-1K and downstrea COCO object detection and instance segmentation, ADE20K semantic segmentation tasks, as shown in Tab. \ref{tbl:ft:imagenet:hyperparams}, Tab. \ref{tbl:ft:od:hyperparams} and Tab. \ref{tbl:ft:semseg:hyperparams}.

\begin{table}[ht]
\centering
\caption{
Hyperparameters for fine-tuning \net~on ImageNet-1K classification.
}
\small
\begin{tabular}{l|c}
\toprule
\multirow{2}[2]{*}{\bf Hyperparameters} & \bf ImageNet-1K\\
\cmidrule(lr){2-2} 
& \bf B/L/XL/H \\

\midrule
Input resolution & $384^2$/$384^2$/$384^2$/$640^2$ \\
Fine-tuning epochs & 30 \\
Warmup epochs & 0 \\
Batch size & 512 \\
Peak learning rate & 5e-5 \\
Layer-wise learning rate decay & 0.9/0.8/0.8/0.8 \\
AdamW momentum & (0.9, 0.999) \\
Weight decay & 1e-8 \\
Learning rate schedule & cosine \\
Head init scale & 0.001\\
Drop path & 0.2/0.3/0.4/0.5 \\
EMA & \xmark/\xmark/\xmark/0.9999 \\
Gradient clipping & 10.0 (norm) \\
\midrule
Label smoothing $\varepsilon$ & 0.1 \\
Data augment & RandAug (9, 0.5) \\
Mixup & \xmark \\
CutMix & \xmark \\
Random erase & 0.25 \\
\bottomrule
\end{tabular}

\label{tbl:ft:imagenet:hyperparams}
\end{table}

\begin{table}[H]
\centering
\small
\caption{
Hyperparameters for fine-tuning \net~on object detection with Cascade Mask R-CNN detector.
}
\begin{tabular}{l|cc}
\toprule
\multirow{2}[2]{*}{\bf Hyperparameters} & \bf IN-1K Pre-trained & \bf IN-22K Pre-trained\\
\cmidrule(lr){2-2} \cmidrule(lr){3-3} 
& \bf \net-T/S/B & \bf \net-B/L\\
\midrule
Fine-tuning epochs & \multicolumn{2}{c}{36} \\
Batch size & \multicolumn{2}{c}{16} \\
Peak learning rate & 2e-4 & 1e-4 \\
Warmup steps & \multicolumn{2}{c}{1500} \\
Layer-wise learning rate decay & 0.85/0.8/0.8 & 0.9/0.8 \\
\midrule
AdamW momentum & \multicolumn{2}{c}{(0.9, 0.999)} \\
Weight decay & \multicolumn{2}{c}{0.05} \\
Drop path & 0.3/0.4/0.4 & 0.5/0.6\\
\bottomrule
\end{tabular}

\label{tbl:ft:od:hyperparams}
\end{table}

\begin{table}[H]
\centering
\small
\caption{
Hyperparameters for fine-tuning \net~on ADE20K semantic segmentation with UperNet segmentation framework.
}
\begin{tabular}{l|cc}
\toprule
\multirow{2}[2]{*}{\bf Hyperparameters} & \bf IN-1K Pre-trained & \bf IN-22K Pre-trained\\
\cmidrule(lr){2-2} \cmidrule(lr){3-3} 
& \bf \net-T/S/B & \bf \net-B/L\\
\midrule
Input resolution & $512^2$ & $640^2$ \\
Fine-tuning steps & \multicolumn{2}{c}{80k} \\
Batch size & \multicolumn{2}{c}{16} \\
Peak learning rate &\multicolumn{2}{c}{4e-5} \\
Warmup steps & \multicolumn{2}{c}{1500} \\
Layer-wise learning rate decay & 1.0 & 0.9 \\
\midrule
AdamW momentum & \multicolumn{2}{c}{(0.9, 0.999)} \\
Weight decay & \multicolumn{2}{c}{0.01} \\
Drop path & \multicolumn{2}{c}{0.3} \\
\bottomrule
\end{tabular}

\label{tbl:ft:semseg:hyperparams}
\end{table}

\subsubsection{Convolution kernel padding trick in down-stream tasks}
According the results shown in Section \ref{sec:ab_ks}, larger kernel convolution perform better especially in down-stream tasks. To save the pre-training cost meanwhile achieve better performance, we pad the small $3\times3$ convolution kernel in pre-trained model weights to larger size then fine-tune in detection and segmentation tasks. Inspired by \emph{Net2net}~\citep{chen2015net2net} method, we pad the pre-trained $3\times3$ kernel in convolution layer with Gaussian initialized values. To protect the pre-trained kernel from being disturbed by the new padded values, we initialize the padded values with 0 mean and extremely small standard deviations (1e-7). 
We use this trick only with our largest model \net-H. We pad the $3\times3$ kernel in pre-trained model to $7\times7$ kernel size in COCO detection task and $13\times13$ in ADE20k sementatic segmentation task, then fine-tune on corresponding dataset to get the final result. In general, the kernel padding trick leads to 0.5$\sim$0.8 $AP_{box}$ improvement and 0.7$\sim$1.0 mIoU improvement for \net-H model.

\section{Visualizations of Feature Disentanglement}
\label{app:vis}

\begin{figure}[t]
    \centering
    \includegraphics[width=\textwidth]{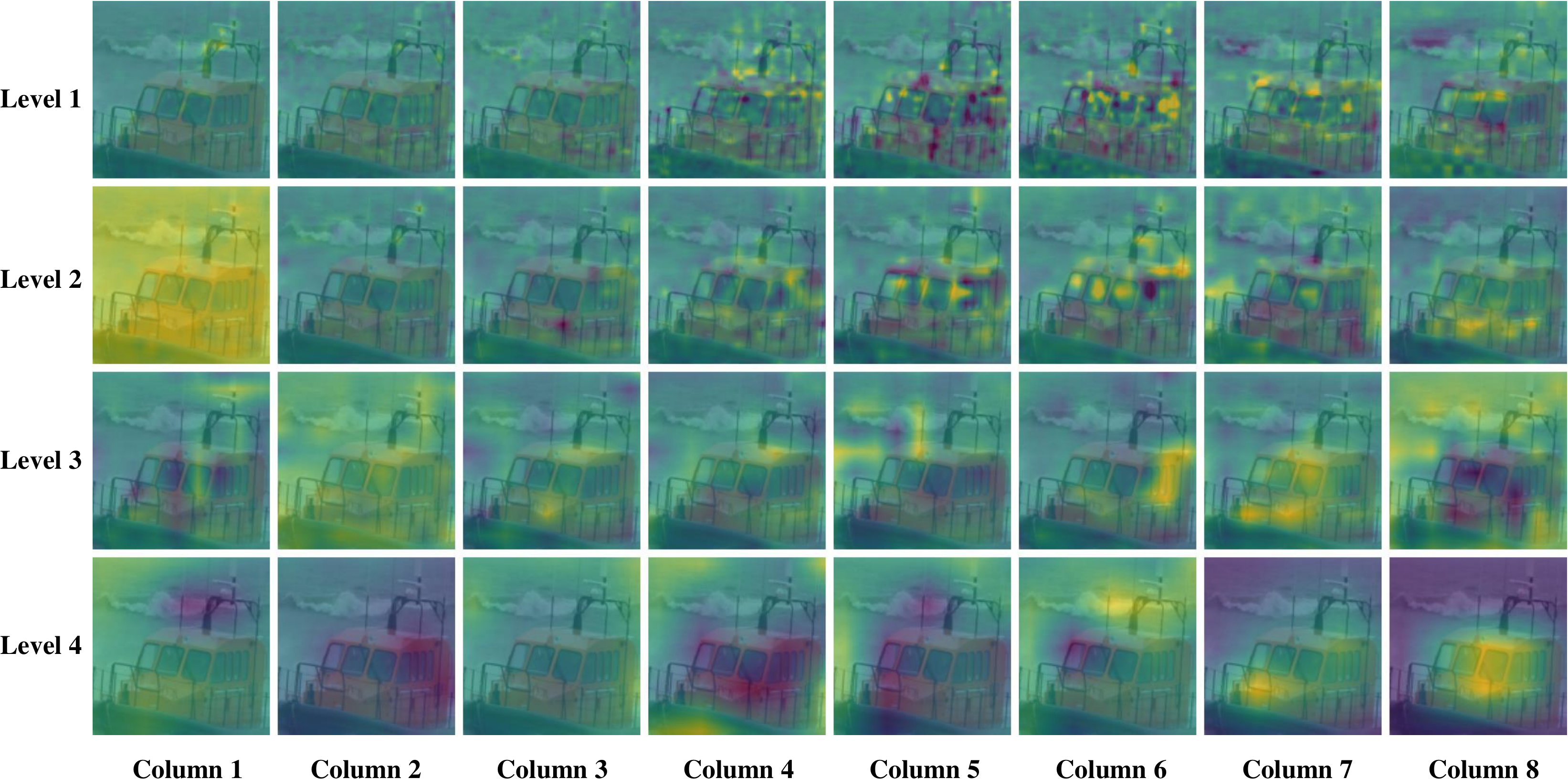}
    \includegraphics[width=\textwidth]{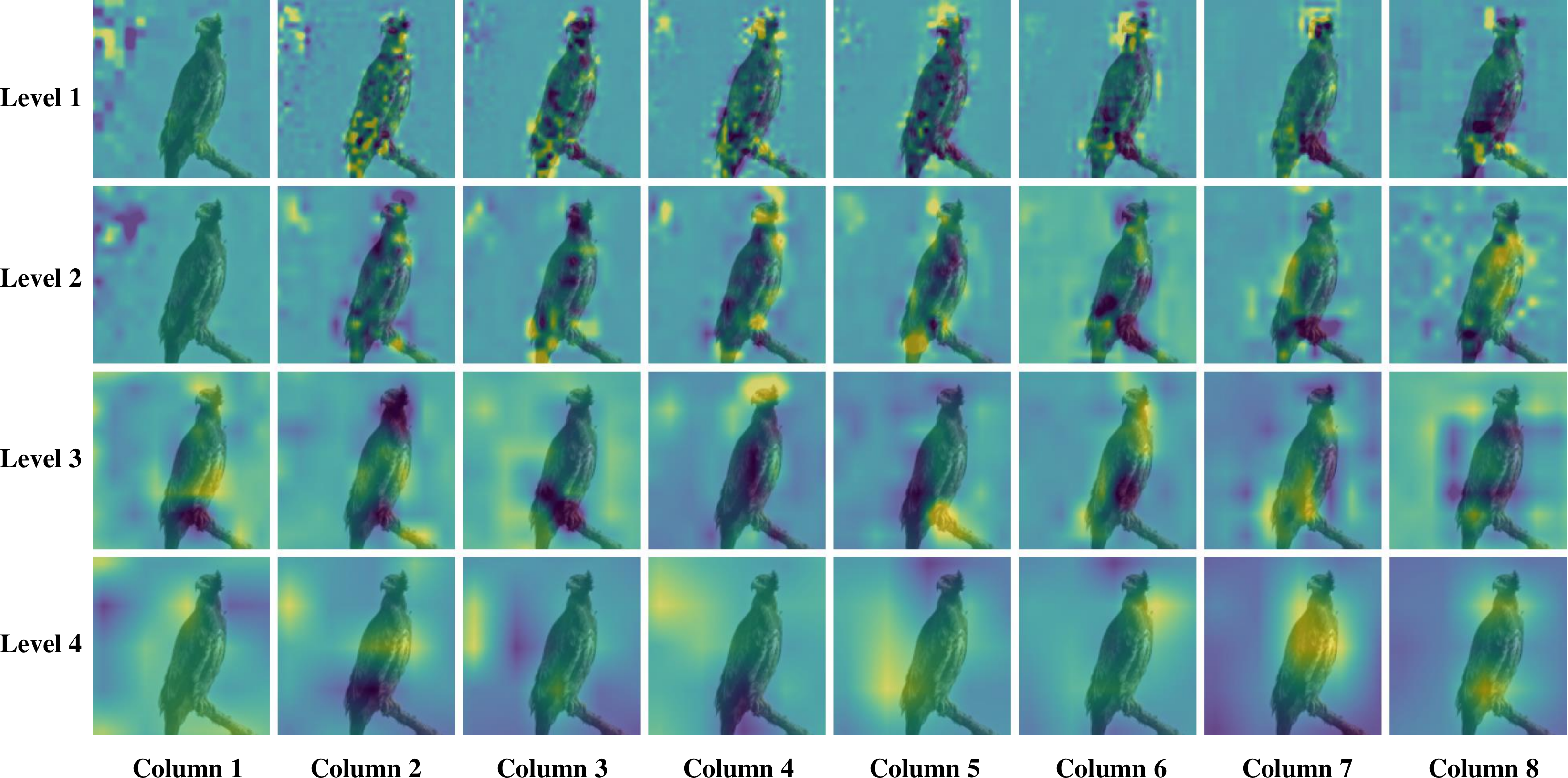}
    \caption{Visualizations of class activation maps using LayerCAM~\citep{jiang2021layercam} for different levels and columns.}
    \label{fig:cam}
\end{figure}

In this section, we show our \net~can disentangle features with stacked columns, which is different from the conventional sequential networks. We use \net-S pre-trained on ImageNet-1K for analysis. 
First, we visualize the class activation maps (CAMs) for outputs of each last layer of a level. We adopt LayerCAM~\citep{jiang2021layercam} technology to generate the CAMs with the predicted classes. Fig. \ref{fig:cam} show the heatmaps of activation. With the levels and columns going deeper, the features focus on the regions with more semantics. The outputs of \net-S are the different levels of last column. These features with high level semantics focus on different parts of the image and the whole part of the object, achieving disentanglement of features for task-relevant and task-irrelevant.

\begin{figure}[t]
    \centering
    \includegraphics[width=\textwidth]{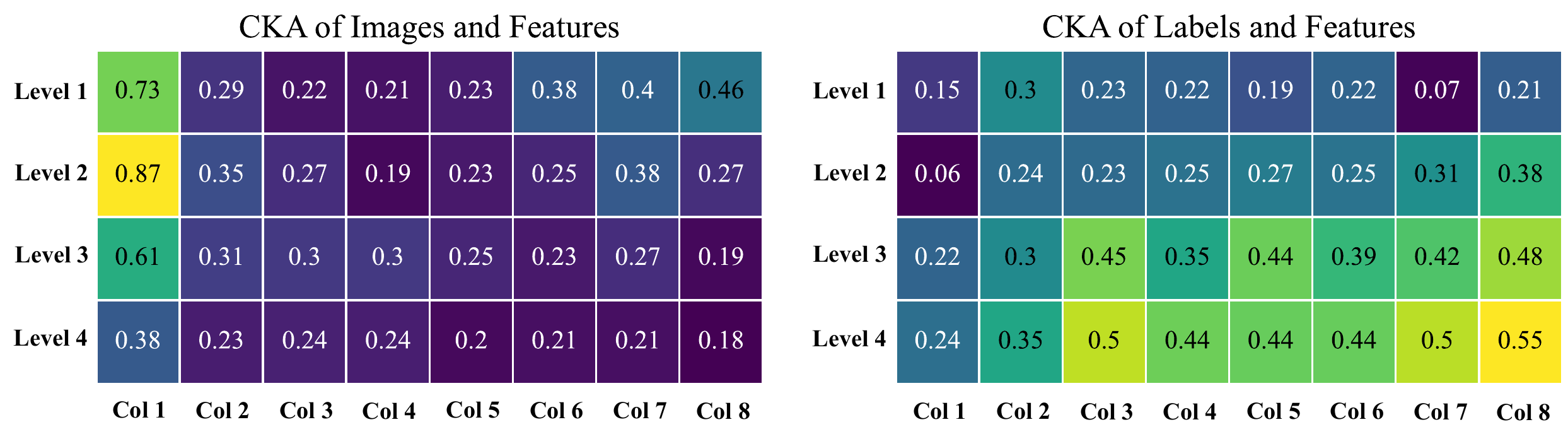}
    \caption{CKA similarities \citep{cka} of features and images/labels for different levels and columns.}
    \label{fig:cka}
\end{figure}

To quantify the disentanglement, we use Centered Kernel Alignment (CKA) similarity metric \citep{cka} to measure the similarity between representations in \net -S. We calculate the CKA similarities between intermediate features in different levels and columns and images or labels of each category in the ImageNet val set. Then we plot the similarities of the category with the highest label similarity in Fig. \ref{fig:cka}. As shown in the figure, the similarities between images and intermediate features are not clearly distinguished at different levels in Column 2-5, while the features with higher levels have lower similarity to the images in Column 6-8. The similarities between labels and intermediate features are also more distinct in higher columns.

\end{document}